\documentclass[smallextended]{svjour3}   
\smartqed

\usepackage{amsmath,amssymb,graphicx}
\usepackage{epsfig}
\usepackage{lscape}
\usepackage{footmisc}
\usepackage{times}
\usepackage{graphicx}
\usepackage{multirow}
\usepackage{lscape}
\usepackage{booktabs}
\usepackage{changepage}
\usepackage{rotating}
\usepackage{array}

\begin{document}
\title{Deep Neural Network Concepts for Background Subtraction: A Systematic Review and Comparative Evaluation}

\author{Thierry Bouwmans, Sajid Javed, Maryam Sultana, Soon Ki Jung}
\institute{Thierry Bouwmans \at
              Lab. MIA, Univ. La Rochelle, France \\
              \email{tbouwman@univ-lr.fr} 
							\and
              Sajid Javed \at
						Dept. of Computer Science, University of Warwick, UK \\
						 \email{s.javed.1@warwick.ac.uk}
             \and     
					    Maryam Sultana \at
						  Dept. of Computer Science and Engineering, Kyungpook National University, Republic of Korea \\
							\email{maryam@vr.knu.ac.kr}
					    \and
              Soon Ki Jung \at
						  Dept. of Computer Science and Engineering, Kyungpook National University, Republic of Korea \\
							\email{skjung@knu.ac.kr}
            }

\date{Received: date / Accepted: date}

\maketitle

\begin{abstract}
Conventional neural networks show a powerful framework for background subtraction in video acquired by static cameras. Indeed, the well-known SOBS method and its variants based on neural networks were the leader methods on the large-scale CDnet 2012 dataset during a long time. Recently, convolutional neural networks which belong to deep learning methods were employed with success for background initialization, foreground detection and deep learned features. Currently, the top current background subtraction methods in CDnet 2014 are based on deep neural networks with a large gap of performance in comparison on the conventional unsupervised approaches based on multi-features or multi-cues strategies. Furthermore, a huge amount of papers was published since 2016 when Braham and Van Droogenbroeck  published their first work on CNN applied to background subtraction providing a regular gain of performance. In this context, we provide the first review of deep neural network concepts in background subtraction for novices and experts in order to analyze this success and to provide further directions. For this, we first surveyed the methods used background initialization, background subtraction and deep learned features. Then, we discuss the adequacy of deep neural networks for background subtraction. Finally, experimental results are presented on the CDnet 2014 dataset.

\end{abstract}

\section{Introduction}
\label{sec:Introduction}
In the last two decades, background subtraction for video taken by static cameras has been one of the most active research topics in computer vision due to a big amount of applications including intelligent surveillance as human activities in public spaces, traffic monitoring and industrial machine vision \cite{Survey-1,Survey-2,Survey-5,Survey-4,Survey-3}. A big variety of models was used for background/foreground separation such as statistical models \cite{P1C2-KDE-2,P1C2-MOG-1000,P1C2-MOG-10,P1C2-MOG-636}, fuzzy models \cite{7701,7700,7702}, subspace learning models \cite{7607,7610,7600}, robust PCA models \cite{RPCA-1010-5,RPCA-1010-4,RPCA-1010-6,RPCA-1010-3,RPCA-1010-2}, and neural networks models \cite{P1C5-417,P1C5-415,P1C5-1}. Similarly as PCA models which renewed interest for this task due to the theoretical advances in robust PCA made in 2009 by Cand\`{e}s et al. \cite{RPCA} after an empty period, neural networks progressively renews interest in this field since 2014 \cite{P1C5-2000} due to the practical advances in deep neural networks which are currently usable due to the availability of large-scale datasets \cite{Dataset-2}\cite{Dataset-1} for the training, and the progress in computational hardware ability\protect\footnotemark[1]. 

\footnotetext[1]{{https://www.nvidia.fr/deep-learning-ai/}}

By looking at the story, Schofield et al. \cite{P1C5-1} were the first authors who used neural networks for background modeling and foreground detection by using a Random Access Memory (RAM) neural networks. But, RAM-NN required that the images represent the background of the scene correctly, and  there is not a background maintenance stage because once the RAM-NN is trained with a single pass of background images, it is not possible to modify this information. In a further work,  Jimenez et al. \cite{P1C5-10} classified each zone of a video frame into three classes of background: static, noisy, and impulsive. The classification is performed with a multilayer Perceptron Neural Network which requires a training set from specific zones of each training frame.  In another work, Tavakkoli \cite{P1C5-20} proposed a neural network approach under the concept of novelty detector. During the training step, the background is divided in blocks. Each block is associated to a Radial Basis Function Neural Network (RBF-NN). Thus, each RBF-NN is trained with samples of the background corresponding to its associated block. The decision of using RBF-NN is because it works like a detector and not a discriminant, generating a close boundary for the known class. RBF-NN methods is able to address dynamic object detection as a single class problem, and to learn the dynamic background. However, it requires a huge amount of samples to represent general background scenarios. In Wang et al. \cite{P1C5-60}, a hybrid probabilistic and "Winner Take All"  (WTA) neural architectures were combined into a single NN model. The algorithm is named Adaptive Background Probabilistic Neural Network (ABPNN) and it is composed of four layers. In the ABPNN model, each pixel is classified as foreground or background according to a conditional probability of being background. This probability is estimated by a Parzen estimation. The foreground regions are further analyzed in order to classify them as a motion or a shadow region. But, ABPNN needs to define specific initial parameter values (specific thresholds values) for each of the analyzed video. In Culibrk et al. \cite{P1C5-70}, a feed-forward neural network is used for background modeling based on an adaptive Bayesian model called Background Neural Network (BNN). The architecture corresponds to a General Regression Neural Network (GRNN), that works like a Bayesian classifier. Although the architecture is proposed as supervised, it can be extended as an unsupervised architecture in the background model domain. The network is composed of three sub-networks:  classification, activation, and replacement. The classifier sub-network maps the features background/foreground of a pixel to a probabilistic density function using the Parzen estimator. The network has two neurons, one of them estimates the probability of being background, and the other neuron computes the probability of being foreground. But, the main disadvantages are that the model is very complex and that it requires of three networks to define if a pixel belongs to the background. In a remarkable work, Maddalena and Petrosino \cite{P1C5-400} proposed a method called Self Organizing Background Subtraction (SOBS) based on a 2D self-organizing neural network architecture preserving pixel spatial relations. The method is considered as nonparametric, multi-modal, recursive and pixel-based. The background is automatically modeled through the neurons weights of the network. Each pixel is represented by a neural map with $n \times n$ weight vectors. The weights vectors of the neurons are initialized with the corresponding color pixel values using the HSV color space.  Once the model is initialized, each new pixel information from a new video frame is compared to its current model to determine if the pixel corresponds to the background or to the foreground. In further works, SOBS was improved in several variants such as Multivalued SOBS \cite{P1C5-405}, SC-SOBS \cite{P1C5-408}, 3dSOBS+ \cite{P1C5-410}, Simplified SOM \cite{P1C5-412}, Neural-Fuzzy SOM \cite{P1C5-414} and MILSOBS \cite{P1C5-421}) which allow this method to be in the leader methods on the CDnet 2012 dataset \cite{Dataset-2} during a long time. SOBS show only interesting performance for stopped object detection \cite{P1C5-403,P1C5-406,P1C5-409}. But, one of the main disadvantages of SOBS based methods is the need to manual adjust at least four parameters. 

Recently, deep learning methods based on Deep Neural Networks (DNNs) with Convolutional Neural Networks (CNNs also called ConvNets) allow to alleviate the disadvantages of these previous approaches based on conventional neural networks \cite{DNN-3000}\cite{DNN-3001}\cite{DNN-3002}. While CNNs existed for a long time, their success and then their use in computer vision was limited during a long period due to the size of the available training sets, the size of the considered networks, and the computational capacity. The breakthrough was made by Krizhevsky et al. \cite{I-2} who used a supervised training of a large network with $8$ layers and millions of parameters on the ImageNet dataset \cite{P1C5-2200-1} with $1$ million training images. Since this work, even larger and deeper networks have been trained with the progress made by the storage for Big Data and by the GPUs for deep learning. For the field of background/foreground separation, DNNs were applied with success 1) for background generation \cite{P1C5-1900,P1C5-2130,P1C5-1910,P1C5-2001,P1C5-2000}, 2) for background subtraction \cite{P1C5-2140,P1C5-2120,P1C5-2100,P1C5-2160,P1C5-2150}, 3) foreground detection enhancement \cite{P1C5-2175}, 4) for ground-truth generation \cite{P1C5-2110}, and 5) for learned deep spatial features \cite{P1C5-2173,P1C5-2171,P1C5-2200,P1C5-2210,P1C5-2010}. More practically, Restricted Boltzman Machine (RBM) was employed by Guo and Qi \cite{P1C5-1900} and Xu et al. \cite{P1C5-1910} for background generation in order to further achieve moving object detection by background subtraction. In a similar manner, Xu et al. \cite{P1C5-2001,P1C5-2000} used deep auto-encoder networks to achieve the same task  while Qu et al. \cite{P1C5-2130} used context-encoder for background initialization. In another approach, Convolutional Neural Networks (CNNs) were employed for background subtraction by Braham and Droogenbroeck \cite{P1C5-2100}, Bautista et al. \cite{P1C5-2120}, and Cinelli \cite{P1C5-2160}. Other authors employed improved CNNs like Cascaded CNNs \cite{P1C5-2110}, deep CNNs \cite{P1C5-2140}, structured CNNs \cite{P1C5-2150} and two stage CNNs \cite{P1C5-2161}. In another way, Zhang et al. \cite{P1C5-2010} used Stacked Denoising Auto-Encoder (SDAE) to learn robust spatial features and modeled the background with density analysis whilst Shafiee et al. \cite{P1C5-2200} employed Neural Reponse Mixture (NeREM) to learn deep features used in the Mixture of Gaussians (MOG) model \cite{P1C2-MOG-10}. Motivations and contributions of this paper can be summarized as follows:
\begin{itemize}
\item Numerous papers were published in the field of background subtraction since the work of Braham et al. in 2016 showing the big interest of deep neural networks in this field. Furthermore, each new method is in the top algorithms on the CDnet 2014 dataset by offering a big gap of performance compared to conventional approaches. In addition, deep neural networks was also employed in background initialization, foreground detection enhancement, ground-truth generation and deep learned features showing its potential in all the field of background subtraction. 
\item In this context, we provide an exhaustive comparative survey regarding DNNs approaches used in the field of background  background initialization, background subtraction, foreground detection and features. For this, we compare them in terms of architecture and performance.  
\end{itemize}
The rest of this paper is as follows. First, we provide in Section \ref{DNN} a short reminder on the different key points in deep neural networks for novices. In Section \ref{BackgroundGeneration}, we review the methods based on deep neural networks for background generation in video.  In Section \ref{BackgroundSubtraction}, we provide the methods based on deep neural networks for background subtraction with a full comparative overview in terms of architecture and challenges. In Section \ref{DeepLearnedFeatures}, deep learned features in this field are surveyed. In addition, we also provide a discussion about the adequacy of deep neural networks for background subtraction. Finally, experimental results are presented on the CDnet 2014 dataset in Section \ref{sec:results}, and concluding remarks are given in Section \ref{sec:Conclusion}. 

\section{Deep Neural Networks: A Short Overview}
\label{DNN}
\subsection{Story Aspects}
DNN recently emerges from a long history of neural networks with two empty periods. Since its beginning, more and more sophisticated concepts and related architectures were developed for neural networks and after for deep neural networks. Full surveys were provided by Schmidhuber \cite{DNN-3000} in 2015, Yi et al. \cite{DNN-3060} in 2016, Liu et al. \cite{DNN-3001} in 2017, and Gu et al. \cite{DNN-3002} in 2018. In addition, a full description of the different DNN concepts are available at the Neural Network Zoo website\protect\footnotemark[2]. Here we briefly summarize the main steps of the DNN's story. DNN begins in 1943 with the threshold logic unit (TLU) \cite{DNN-1}. In further works, Rosenblatt  \cite{DNN-2} designed the first perceptron in 1957 whilst Widrow \cite{DNN-3}\cite{DNN-3-1} developed the Adaptive Linear Neuron (ADALINE) in 1962. This first generation of neural networks are fundamentally limited in what they can learn to do. During the 1970s (first empty period), research focused more on XOR problem. The next period concerns the emergence of more advanced neural networks like multilayer back-propagation neural networks, Convolutional Neural Networks (CNNs), and Long Short-Term Memory (LSTMs) for Recurrent Neural Networks (RNNs) \cite{DNN-6}. This second generation of neural networks mostly used back-propagation of the error signal to get derivatives for learning. After 1995 until 2006 (second empty period), research focused more Support Vector Machine (SVM) which is a very clever type of perceptron developed by Vapnik et al. \cite{DNN-7}. Thus, many researchers abandoned neural networks research with multiple adaptive hidden layers because SVM worked better with less computational time requirements and training. With the progress of GPU and the storage of Big Data, DNN regains attention and developments with new deep learning concepts such as a) Deep Belief Networks \cite{RBM-2}\cite{DNN-10} in 2006 and b) Generative Adversarial Networks (GANs) \cite{GAN-1}\cite{GAN-2}in 2014. Liu et al. \cite{DNN-3001} classified  the deep neural network architectures in the following categories: restricted Boltzmann machines (RBMs), deep belief networks (DBNs), autoencoders (AEs) network and deep Convolutional Neural Network (CNNs). In addition, deep probabilistic neural networks \cite{DNN-5000}, deep fuzzy neural networks \cite{DNN-6000}\cite{DNN-6001} and Generative Adversarial Networks (GANs) \cite{GAN-1}\cite{GAN-2} can also be considered as other categories. Applications of these deep learning architecture are mainly in speech recognition, computer vision and pattern recognition \cite{DNN-3001}.In this context, DeepNets architectures for specific applications have emerged such as the following well-known architecture: AlexNet developed by Krizhevsky et al. \cite{I-2} for image classification in 2012, VGG-Net designed by Simonyan and Zisserman \cite{P1C5-2150-2} for large-scale image recognition in 2015, U-Net \cite{P1C5-2164-1} developed by Ronneberger et al. \cite{P1C5-2164-1} for biomedical image segmentation in 2015, GoogLeNet with inception neural network introduced by Szegedy et al. \cite{P1C5-2170-1} for computer vision in 2015, and Microsoft Residual Network (ResNet) designed by He et al. \cite{P1C5-2100-2} for image recognition in 2016. Thus, all the current architectures were designed for a target application like speech recognition  \cite{DNN-3051}, computer vision \cite{DNN-3050} and pattern recognition \cite{DNN-3001} which its specific features giving very impressive performance in comparison on the previous state-of-art methods based on GMM and graph-cut as in the problem of foreground detection/segmentation/localization. 

\footnotetext[2]{{http://www.asimovinstitute.org/neural-network-zoo/}}

\subsection{Features Aspects}
As seen in the previous part, DNNs are determined by their architecture that becomes more and more sophisticated over time. Practically, an architecture consists of different layers classified as input layer, hidden layer and output layer. Each layers contains a number of neurons that are activated or not following an activation function. This activation function can be viewed as the mapping of the input to the output via a non-linear transform function at each node. In literature, different activation functions can be found as the sigmoid function \cite{DNN-3030}, Rectified Linear Unit (ReLU) \cite{DNN-3040}, and Probabilistic ReLU (PReLU) \cite{P1C5-2165-3}. Once the architecture is determined and the activation function is chosen, the DNN need to be trained using a large-scale dataset such as ImageNet dataset \cite{I-2}, CIFAR-10 dataset and ILSVRC 2015 dataset for classification tasks. For this, the architecture is exposed to the training dataset to learn the weights of each neurons in each layer. The parameters are learned via a cost function that are minimized on the desired output and the predicted one. The most common method for training is the back-propagation. Usually, the gradient of the error function computed on the correct output and the predicted one is propagated back to the beginning of the network in order to update its parameters. For this, it requires a gradient descent algorithm. Batch normalization which normalizes mini-batches can also be used to accelerate learning because it employs higher learning rates, and also regularizes the learning. For vocabulary, an epoch is a complete pass through a given dataset, and thus is the number of time where the neural network has been exposed to every record of the dataset once. An epoch is not an iteration which corresponds to one update of the neural net models parameters. Many iterations can occur before an epoch is over. Epoch and iteration are only identical if the parameters are updated once for each pass through the whole dataset. 

\subsection{Theoretical Aspects}
Theoretical aspects concern mainly the understanding and the provability of DNNs \cite{DNN-103,DNN-100,DNN-101,DNN-102}, but also their properties in presence of adversarial perturbations \cite{DNN-9000,GAN-4,GAN-5,GAN-10,GAN-11,GAN-3,GAN-21}, and their robustness in presence of noisy labels \cite{GAN-20}. For this, the principle key features to design DNNs need to be mathematically investigated as follows \cite{DNN-100,DNN-101}:
\begin{itemize}
\item \textbf{Architecture:} The number, the size and the type of the layers are key characteristics of an architecture as well as  the classes of functions that can be approximated by a feed-forward neural network. The key issue is how the chosen architecture impact expressiveness. 
\item \textbf{Optimization:} It concerns the way to train the DNNs. This issue contains two aspects which are the datasets used for the training, and mostly the algorithm to optimize the network. The problem is generally non-convex, and following the appearance of the error surface how to guarantee the optimality and when does descent gradient succeed? Is "the local minima are global property" hold for deep nonlinear networks? 
\item \textbf{Generalization:} How well do DNNs generalize? How should DNNs be regularized? How to prevent under and over fitting? 
\end{itemize}
Both architecture and optimization can impact generalization \cite{DNN-103,DNN-100,DNN-101,DNN-102}. Furthermore, several architectures are easier to optimize thanothers \cite{DNN-100,DNN-101}. First replies about the global optimality can be found in Yun et al \cite{DNN-102}. In addition, Wang et al. \cite{DNN-3020} show that deep neural networks can be better understood by utilizing the knowledge obtained by the visualization of the output images obtained at each layers. Other authors provided either a theoretical analysis or visualizing analysis in a context of an application. For example, Basu et al. \cite{DNN-3010} published a theoretical analysis for texture classification whilst Minematsu et al. \cite{P1C5-2163,P1C5-2163-1} provided a visualizing analysis for background subtraction. Despite these first valuable investigation, the understanding of DNNs remains still shallows. Nevertheless, DNNs have been applied with success in many computer vision applications gaining a big gap of performance. This success is intuitively due to the following reasons: 1) features are learned rather than manual hand-crafted, 2) more layers capture more invariance, 3) more data allow a deeper training, 4) more computing CPU, 5) better regularization (Dropout \cite{DNN-110}) and 6) new non-linearity (max-pooling, ReLU \cite{DNN-120}). 

\subsection{Implementation Aspects}
For software implementation, many libraries for the development in different programming languages are available to implement DNNs. The most known libraries are Caffe \cite{I-1}, MatConvNet \cite{I-3} from Matlab, Microsoft Cognitive Toolkit (CNTK), TensorFlow \cite{I-10},  Theano \protect\footnotemark[3] and Torch \protect\footnotemark[4]. All these software support interfaces of C, C++ and/or Python for quick development. For a full list, the reader are referred to go on the deeplearning.net\protect\footnotemark[5] website. There is also a Deep Learning library for Java (DL4J\protect\footnotemark[6]). For hardware implementation and optimization, there are several designed GPUs from NVIDIA with dedicated SDKs\protect\footnotemark[7]. For example, the deep learning GPU Training System (DIGITS\protect\footnotemark[8])  provides fast training of DNNs for computer vision applications like image classification, segmentation and object detection tasks whilst NVIDIA Jetson is designed for embedded systems. For NVIDIA Volta GPUs, TensorRT protect\footnotemark[9] allow to optimize deep learning inference and runtime. It also allows to deploy trained neural networks for inference to hyper-scale data centers or embedded. Deep neural network accelerator based on FPGA also existed \cite{DNN-3070}.\\

\footnotetext[3]{{http://deeplearning.net/software/theano/}}
\footnotetext[4]{{http://torch.ch/}}
\footnotetext[5]{{http://deeplearning.net/software-links/}}
\footnotetext[6]{{https://deeplearning4j.org/}}
\footnotetext[7]{{https://developer.nvidia.com/deep-learning-software}}
\footnotetext[8]{{https://developer.nvidia.com/digits}}
\footnotetext[9]{{https://developer.nvidia.com/tensorrt}}

In the following sections, we survey all the previous DNN approaches used in background/foreground separation steps by comparing their advantages and disadvantages as well as their performance on the CDnet 2014 dataset. 

\section{Background Generation}
\label{BackgroundGeneration}
Background generation \cite{BI-1,BI-3,BI-2} (also called background initialization \cite{BI-11,BI-13} \\
\cite{BI-12,BI-14}, background estimation \cite{BI-20,BI-7}, and background extraction \cite{BI-30}) regards the initialization of the background. Generally, the model is often initialized using the first frame or a background model over a set of training frames which contain or do not contain foreground objects. This background model can be the temporal average or the the temporal median. But, it is impossible in several environments due to bootstrapping and then it needs a sophisticated model to construct this first image. The top algorithms on the SBMnet dataset are the algorithms named Motion-assisted Spatio-temporal Clustering of Low-rank (MSCL) \cite{BI-10} and LaBGen \cite{BI-5,BI-5-1,BI-5-2} that are based on robust PCA \cite{Survey-5,Survey-4} and the robust estimation of the median, respectively. Practically, the main challenge is to obtain a first background model when more than half of the training contains foreground objects. This learning process can be done off-line and so the algorithm can be a batch one. Thus, deep neural networks are suitable for this task and several DNN methods have been recently used in this field. We have classified them in the following categories and Table \ref{BI-Overview} shows an overview of these methods. In addition, the list of publications is available at the Background Subtraction Website\protect\footnotemark[10] and is regularly updated.\\

\begin{table}
\begin{tabular}{|l|l|l|} 
\hline
\scriptsize{Categories} &\scriptsize{Methods} &\scriptsize{Authors - Dates}\\
\hline
\hline
\multirow{4}{*}{\scriptsize{Restricted Boltzmann Machines}}
& \scriptsize{Partially-Sparse RBM (PS-RBM)}       & \scriptsize{Guo and Qi \cite{P1C5-1900} (2013)}    \\
& \scriptsize{Temp. Adaptive RBM (TARBM)}          & \scriptsize{Xu et al. \cite{P1C5-1910} (2015)}     \\
& \scriptsize{Gaussian-Bernoulli RBM}              & \scriptsize{Sheri et al. \cite{P1C5-1920} (2018)}  \\
& \scriptsize{RBM (PTZ Cameras)}                   & \scriptsize{Rafique et al. \cite{1} (2014)}        \\
\hline 
\multirow{3}{*}{\scriptsize{Deep Auto-encoders Networks}} 
& \scriptsize{Deep Auto-encoder Networks (DAN)}           & \scriptsize{Xu et al. \cite{P1C5-2000} (2014)}   \\
& \scriptsize{DAN with Adaptive Tolerance Measure }       & \scriptsize{Xu et al. \cite{P1C5-2001}  (2014)}  \\
& \scriptsize{Encoder-Decoder CNN (ED-CNN)}               & \scriptsize{Qu et al. \cite{P1C5-2130} (2016)}   \\
\hline
\multirow{2}{*}{\scriptsize{Convolutional Neural Networks}}
& \scriptsize{FC-Flownet}                                 & \scriptsize{Halfaoui et al. \cite{BI-7} (2016)}  \\
& \scriptsize{BM-Unet}                                    & \scriptsize{Tao et al. \cite{P1C5-2164} (2017)}  \\
\hline 
\multirow{1}{*}{\scriptsize{Generative Adversarial Networks}} 
& \scriptsize{Deep Context Prediction (DCP)}           & \scriptsize{Sultana et al. \cite{P1C5-2190} (2018)} \\   
& \scriptsize{ForeGAN-RGBD}                            & \scriptsize{Sultana et al. \cite{P1C5-2190-1} (2018)}  \\   

\hline
\end{tabular}
\caption{Deep Neural Networks in Background Generation: An Overview} \centering
\label{BI-Overview}
\end{table}

\footnotetext[10]{{https://sites.google.com/site/backgroundsubtraction/background-initialization/neural-networks}}

\subsection{Restricted Boltzmann Machines (RBMs)}
Guo and Qi \cite{P1C5-1900} were the first authors who applied Restricted Boltzmann Machine (RBM) to background generation by using a Partially-Sparse RBM (PS-RBM) framework in order to detect moving objects by background subtraction. This framework models the image as the integration of RBM weights. By introducing a sparsity target, the learning process alleviate the tendency of growth in weights. Once the sparse constraints are added to the objective function, the hidden units only keep active in a rather small portion on the specific training data. In this context, Guo and Qi \cite{P1C5-1900} proposed a controlled redundancy technique, that allow the hidden units to learn the distinctive features as sparse as possible, meanwhile, the redundant part rapidly learns the similar information to reduce the total error. The PS-RBM provides accurate background modeling even in dynamic and noisy environments. Practically, PS-RBM provided similar results than DPGMM \cite{P2C1-11}, KDE \cite{P1C2-KDE-2}, KNN \cite{P1C2-KDE-30}, and SOBS \cite{P1C5-400} methods on the CDnet 2012 dataset. 

In a further work, Xu et al. \cite{P1C5-1910} proposed a Temporally Adaptive RBM (TARBM) background subtraction to take into account the spatial coherence by exploiting possible hidden correlations among pixels while exploiting the temporal coherence too.  As a result, the augmented temporally adaptive model can generate more stable background given noisy inputs and adapt quickly to the changes in background while keeping all the advantages of PS-RBM including exact inference and effective learning procedure. TARBM outperforms the standard RBM, and it is robust in presence of dynamic background and illumination changes. 

Sheri et al. \cite{P1C5-1920} employed a Gaussian-Bernoulli restricted Boltzmann machine (GRBM) which is different from the ordinary restricted Boltzmann machine (RBM) by using real numbers as inputs. This network results in a constrained mixture of Gaussians, which is one of the most widely used techniques to solve the background subtraction problem. Then, GRBM easy learn the variance of pixel values and takes the advantage of the generative model paradigm of the RBM.

In the case of PTZ cameras, Rafique et al. \cite{1} modeled the background scene by using RBM. The generative modeling paradigm of RBM gives an extensive and nonparametric background learning framework. Then, RBM was trained with one step contrastive divergence. \\

\subsection{Deep Auto Encoder Networks (DAE)}
Xu et al. \cite{P1C5-2000} designed a background generation method based on two auto-encoder neural net-works. First, the approximate background images are computed via an auto-encoder network called Reconstruction Network (RN) from the current video frames. Second, the background model is learned based on these background images with another auto-encoder network called Background Network (BN). In addition, the background model is updated on-line to incorporate more training samples over time. Experimental results on the I2R dataset \cite{P6C2-Dataset-10} show that DAN outperforms MOG \cite{P1C2-MOG-10}, Dynamic Group Sparsity (DGS) \cite{P3C4-DGS-10}, Robust Dictionary Learning (RDL) \cite{P3C4-DL-23} and Online RDL (ORDL) \cite{P3C4-DL-26}. In a further work, Xu et al. \cite{P1C5-2001} improved this method by using an Adaptive Tolerance Measure Thus, DAN-ATM can handle large variations of dynamic background more efficiently than DAN. Experimental results on the I2R dataset \cite{P6C2-Dataset-10} confirm this gap of performance.

Qu et al. \cite{P1C5-2130} employed a context-encoder network for a motion-based background generation method by removing the moving foreground objects and learning the feature. After removing the foreground, a context-encoder is also used to predict the missing pixels of the “empty” region, and to generate a background model of each frame. The architecture is based on the AlexNet architecture that produces a latent feature representation of input image samples with “empty” regions. The decoder has five up convolutional layers, and uses the feature representation to fill the missing regions of the input samples. The encoder and the decoder are connected through a channel-wise fully connected layer. It allows information to be propagated within activations of each feature map. Experiments provided by Qu et al.  \cite{P1C5-2130} are limited but convincing.

\subsection{FC-FlowNet}
Halfaoui et al. \cite{BI-7} employed a CNN architecture for background estimation which can provide a background image with just a small set of frames containing foreground objects. The CNN is trained estimate background patches and then it is followed by a post-processing step to obtain the final background image. The architecture is based on FlownNetSimple \cite{BI-7-1} which is a two-stage architecture developed for the prediction of the optical flow motion vectors. The first stage is a contractive stage whilst the a second one is a refinement stage. The contractive stage is a succession of convolutional layers. This rather generic stage extracts high level abstractions of the stacked input images, and forwards the gained feature maps to the up convolutional refinement stage, in order to enhance the coarse-to-fine transformations. Halfaoui et al. \cite{BI-7} adapted this architecture by providing a  Fully-concatenated version called FCFlowNet. Experimental results \cite{BI-7} on the SBMC 2016 dataset\protect\footnotemark[11]
demonstrates robustness against very short or long sequences, dynamic background, illumination changes and intermittent object motion. \\

\footnotetext[11]{{http://pione.dinf.usherbrooke.ca/sbmc2016/}}

\subsubsection{U-Net}
Tao et al. \cite{P1C5-2164} proposed an unsupervised deep learning model for Background Modeling called BM-Unet. This method is based on the generative architecture U-Net \cite{P1C5-2164-1} which for a given frame (input) provides the corresponding background image (output) with a probabilistic heat map of the color values. In addition, this method learns parameters automatically and uses  intensity differences and optical flow features in addition of color features to tackle camera jitter and quick illumination changes Besides, BM-Unet can be applied on a new video sequence without the need of re-training. Practically, Tao et al. [45] proposed two algorithms named Baseline BM-Unet and Augmented BM-Unet that can handle static background and background with illumination changes and camera jitter, respectively. The BM-Unet is based on the so called “guide” features which are used to guide the network to generate the background corresponding to the target frame. Experimental results \cite{P1C5-2164} on the SBMnet dataset\protect\footnotemark[12] \cite{BI-3} demonstrate promising results over neural networks methods (BEWiS \cite{P1C5-1602}, BE-AAPSA \cite{BI-6}, and FC-FlowNet \cite{BI-7}), and state-of-the-art methods (Photomontage \cite{BI-4}, LabGen-P \cite{BI-5}). \\

\footnotetext[12]{{http://scenebackgroundmodeling.net/}}

\subsection{Generative Adversarial Networks (GANs)}
Generative Adversarial Networks (GAN) have been a breakthrough in machine learning. Introduced in 2014, GAN \cite{GAN-1}\cite{GAN-2} provide a powerful framework for using unlabeled data to train machine learning models, rising as one of the most promising paradigms for unsupervised learning. Based on GAN, Sultana et al. \cite{P1C5-2190} designed an unsupervised Deep Context Prediction (DCP) for background initialization in the context of background/foreground separation. Practically, DCP is an unsupervised visual feature learning hybrid GAN based on context prediction. It is followed by a semantic inpainting network for texture optimization. Sultana et al. \cite{P1C5-2190} trained the context prediction model addition-
ally with scene-specific data in terms of patches of size $128 \times 128$ for $3$ epochs. The texture optimization is done with
VGG?19 network pre-trained on ImageNet \cite{P1C5-2200-1} for classification. Then, the frame selection for inpainting the background is done by summation of pixel values in the forward frame difference technique. If the sum of difference pixels is small, then current frame is selected. Experimental results on the SBM.net dataset \cite{BI-3}  show that DCP achieved an average gray level error to be $8.724$ which is minimum among all the compared low-rank methods, that are RFSA \cite{RPCA-1033}, GRASTA \cite{RPCA-27}, GOSUS \cite{RPCA-85}, SSGoDec \cite{RPCA-6}, and DECOLOR \cite{RPCA-25}. In a further work,  Sultana et al. \cite{P1C5-2190-1} extended this method to RGB-D videos by separately training two DCPs: one for RGB videos and one for depth videos. Then, each generated background sample is then subtracted from the given test sample to detect foreground objects either in RGB or in depth. Finally, the final foreground mask is obtained by combining the two foreground masks with a logical AND. Experiments on the SBM-
RGBD\protect\footnotemark[13] dataset \cite{P6C2-Dataset-1026} show that ForeGAN-RGBD model outperforms cwisardH+ \cite{P1C5-1604}, RGB-SOBS \cite{P1C5-408}, and SRPCA \cite{RPCA-1010-13} with an average F-Measure of $0.8966$. \\

\footnotetext[13]{{http://rgbd2017.na.icar.cnr.it/SBM-RGBDdataset.html}}

\begin{table}
\scalebox{0.70}{
\begin{tabular}{|l|l|l|} 
\hline
\scriptsize{Categories} &\scriptsize{Methods} &\scriptsize{Authors - Dates}\\
\hline
\hline
\multirow{3}{*}{\scriptsize{Convolutional Neural Networks}} 
& \scriptsize{CNN (ConvNets)}                          & \scriptsize{Braham and Van Droogenbroeck \cite{P1C5-2100} (2016)}       \\
& \scriptsize{CNN (ConvNets)}                          & \scriptsize{Bautista et al. \cite{P1C5-2120} (2016)}  \\
& \scriptsize{CNN (ConvNets) (Analysis) (2)}           & \scriptsize{Minematsu et al. \cite{P1C5-2163} (2017)} \\
& \scriptsize{CNN (Pedestrian Detection)} 	           & \scriptsize{Yan et al. \cite{P1C5-2180} (2018)}       \\
& \scriptsize{CNN (GoogLeNet)} 	                       & \scriptsize{Weinstein  \cite{P1C5-2181} (2018)}       \\
& \scriptsize{CNN (RPoTP feature)} 	                   & \scriptsize{Zhao et al. \cite{P1C5-2182} (2018)}     \\
& \scriptsize{CNN (Depth feature)} 	                   & \scriptsize{Wang et al. \cite{P1C5-2183} (2018)}     \\
\hline  
\multirow{6}{*}{\scriptsize{Multi-scale and Cascaded CNN}} 
& \scriptsize{Cascaded CNN (Ground-Truth)}           & \scriptsize{Wang et al. \cite{P1C5-2162} (2016)}       \\
& \scriptsize{FgSegNet-M}                            & \scriptsize{Lim and Keles  \cite{P1C5-2168} (2018)}    \\
& \scriptsize{FgSegNet-S}                            & \scriptsize{Lim and Keles  \cite{P1C5-21680} (2018)}   \\
& \scriptsize{FgSegNet-V2}                           & \scriptsize{Lim et al. \cite{P1C5-21681} (2018)}       \\
& \scriptsize{MCSS}                                  & \scriptsize{Liao et al. \cite{P1C5-2179} (2018)}       \\
& \scriptsize{Guided Multi-scale CNN}                & \scriptsize{Liang et al. \cite{P1C5-2177} (2018)}       \\ 
\hline
\multirow{6}{*}{\scriptsize{Fully CNNs}} 
& \scriptsize{Basic Fully CNN}                                     & \scriptsize{Cinelli \cite{P1C5-2160} (2017)}      \\
& \scriptsize{Basic Fully CNN}                                     & \scriptsize{Yang et al. \cite{P1C5-2165} (2017)}  \\
& \scriptsize{Multiview recep. field FCN (MV-FCN)}                 & \scriptsize{Akilan et al.\cite{P1C5-2170} (2018)} \\
& \scriptsize{Multiscale Fully CNN (MFCN)}                         & \scriptsize{Zeng and Zhu \cite{P1C5-2174} (2018)} \\
& \scriptsize{CNN-SFC (Foreground Masks)}                                             & \scriptsize{Zeng et al. \cite{P1C5-2175} (2018)}  \\
& \scriptsize{Fully Conv. Semantic Net. (FCSN)}                    & \scriptsize{Lin et al. \cite{P1C5-2178} (2018)}    \\
\hline
\multirow{4}{*}{\scriptsize{Deep CNN}} 
& \scriptsize{Deep CNNs}                                            & \scriptsize{Babaee et al. \cite{P1C5-2140} (2017)}  \\
& \scriptsize{TCNN/Joint TCNN}                                     & \scriptsize{Zhao et al \cite{P1C5-2161} (2017)}     \\
& \scriptsize{Adaptive deep CNN (ADCNN)}                           & \scriptsize{Li et al. \cite{P1C5-2162} (2018)}      \\
& \scriptsize{SFEN}                                                & \scriptsize{Chen et al. \cite{P1C5-2166} (2018)}    \\
\hline
\multirow{1}{*}{\scriptsize{Structured CNN}} 
& \scriptsize{Struct CNNs}                                          & \scriptsize{Lim et al. \cite{P1C5-2150} (2017)}  \\
\hline
\multirow{3}{*}{\scriptsize{3D CNNs}} 
& \scriptsize{3D-CNNs}                                          & \scriptsize{Sakkos et al. \cite{P1C5-2167}  (2017)}  \\
& \scriptsize{STA-3D ConvNets (ReMoteNet)}                     & \scriptsize{Yu et al. \cite{P1C5-2169} (2017)}  \\
& \scriptsize{3D Atrous CNN (ConvLSTM)}                        & \scriptsize{Hu et al. \cite{P1C5-2176} (2018)}  \\
\hline
\multirow{4}{*}{\scriptsize{Generative Adversarial Networks}} 
& \scriptsize{BScGAN}                                          & \scriptsize{Bakkay et al. \cite{P1C5-2191} (2018)}   \\
& \scriptsize{Bayesian GAN (BGAN)}                             & \scriptsize{Zheng et al. \cite{P1C5-2192} (2018)}    \\
& \scriptsize{Bayesian Parallel Vision GAN (BPVGAN)}           & \scriptsize{Zheng et al. \cite{P1C5-2193} (2018)}     \\
& \scriptsize{Neural Unsupervised Moving Object Detection (NUMOD)} & \scriptsize{Bahri et al. \cite{P1C5-2194} (2018)}  \\
\hline
\end{tabular}}
\caption{Deep Neural Networks in Background Subtraction: An Overview} \centering
\label{BGS-Overview1}
\end{table}

\begin{table}
\scalebox{0.65}{
\begin{tabular}{|l|l|l|l|l|l|l|l|l|} 
\hline
\scriptsize{Methods} &\scriptsize{Input} &\scriptsize{Output} &\scriptsize{Architecture} &\scriptsize{Additional} &\scriptsize{Activation} &\scriptsize{Conv.} &\scriptsize{Fully Conv.} &\scriptsize{Implementation} \\
\scriptsize{} &\scriptsize{} &\scriptsize{} &\scriptsize{Encoder/Decoder} &\scriptsize{Architecture} &\scriptsize{Function} &\scriptsize{Layers} &\scriptsize{} &\scriptsize{Framework}  \\
\hline
\hline
\scriptsize{\textbf{Basic CNNs}} &\scriptsize{} &\scriptsize{} &\scriptsize{} &\scriptsize{} &\scriptsize{} &\scriptsize{} &\scriptsize{} &\scriptsize{} \\
\scriptsize{ConvNets \cite{P1C5-2100}} &\scriptsize{Backg. (Median)} &\scriptsize{Foreground} &\scriptsize{LeNet-5 \cite{P1C5-2100-1}} &\scriptsize{-} &\scriptsize{ReLU/Sigm.} &\scriptsize{2} &\scriptsize{1} &\scriptsize{-} \\
\scriptsize{} &\scriptsize{Current Image} &\scriptsize{} &\scriptsize{} &\scriptsize{} &\scriptsize{} &\scriptsize{} &\scriptsize{} &\scriptsize{} \\
\scriptsize{Basic CNNs \cite{P1C5-2162}} &\scriptsize{Current Image} &\scriptsize{Foreground} &\scriptsize{CNN-1} &\scriptsize{-} &\scriptsize{ReLU/Sigm.} &\scriptsize{4} &\scriptsize{2} &\scriptsize{Caffe \cite{I-1}/MatConvNet \cite{I-3}} \\
\scriptsize{Basic CNNs \cite{P1C5-2180}} &\scriptsize{Backg. Visible (Median)} &\scriptsize{GT} &\scriptsize{CNN} &\scriptsize{-} &\scriptsize{ReLU/Sigm.} &\scriptsize{4} &\scriptsize{-} &\scriptsize{-} \\
\scriptsize{} &\scriptsize{Backg. Thermal (Median)} &\scriptsize{} &\scriptsize{} &\scriptsize{} &\scriptsize{} &\scriptsize{} &\scriptsize{} &\scriptsize{} \\
\scriptsize{} &\scriptsize{Current Image (Visible)} &\scriptsize{} &\scriptsize{} &\scriptsize{} &\scriptsize{} &\scriptsize{} &\scriptsize{} &\scriptsize{} \\
\scriptsize{} &\scriptsize{Current Image (Thermal)} &\scriptsize{} &\scriptsize{} &\scriptsize{} &\scriptsize{} &\scriptsize{} &\scriptsize{} &\scriptsize{} \\
\scriptsize{Basic CNNs \cite{P1C5-2181}} &\scriptsize{Backg. (Median)} &\scriptsize{Foreground} &\scriptsize{GoogLeNet \cite{P1C5-2170-1}} &\scriptsize{-}&\scriptsize{ReLU/Sigm.} &\scriptsize{-} &\scriptsize{-} &\scriptsize{Tensorflow \cite{I-10}} \\
\scriptsize{} &\scriptsize{Current Image} &\scriptsize{(Bound. Box)} &\scriptsize{} &\scriptsize{} &\scriptsize{} &\scriptsize{} &\scriptsize{} &\scriptsize{} \\
\scriptsize{Basic CNNs \cite{P1C5-2182}} &\scriptsize{Current Image (RPoTP)} &\scriptsize{Foreground} &\scriptsize{CNN} &\scriptsize{-}&\scriptsize{ReLU} &\scriptsize{-} &\scriptsize{1} &\scriptsize{-} \\
\scriptsize{Basic CNNs \cite{P1C5-2183}} &\scriptsize{Background Image (Average) (Depth)} &\scriptsize{Foreground} &\scriptsize{CNN} &\scriptsize{(MLP)}&\scriptsize{ReLU/Sigmoid} &\scriptsize{3} &\scriptsize{3} &\scriptsize{-} \\
\scriptsize{} &\scriptsize{Current Image (Depth)} &\scriptsize{} &\scriptsize{} &\scriptsize{-}&\scriptsize{-} &\scriptsize{-} &\scriptsize{-} &\scriptsize{-} \\
\hline
\scriptsize{\textbf{Multi-scale and Cascaded CNNs}} &\scriptsize{} &\scriptsize{} &\scriptsize{} &\scriptsize{} &\scriptsize{} &\scriptsize{} &\scriptsize{} &\scriptsize{} \\
\scriptsize{Multi-scale CNNs \cite{P1C5-2162}} &\scriptsize{Current Image} &\scriptsize{GT} &\scriptsize{CNN-1} &\scriptsize{-} &\scriptsize{ReLU/Sigm.} &\scriptsize{-} &\scriptsize{-} &\scriptsize{Caffe \cite{I-1}/MatConvNet \cite{I-3}} \\
\scriptsize{Cascaded CNNs \cite{P1C5-2162}} &\scriptsize{Current Image} &\scriptsize{GT} &\scriptsize{CNN-1} &\scriptsize{CNN-2} &\scriptsize{ReLU/Sigm.} &\scriptsize{-} &\scriptsize{-} &\scriptsize{Caffe \cite{I-1}/MatConvNet \cite{I-3}} \\
\scriptsize{FgSegNet-M \cite{P1C5-2168}} &\scriptsize{Current Image} &\scriptsize{Foreground} &\scriptsize{VGG-16 \cite{P1C5-2150-2}} &\scriptsize{TCNN} &\scriptsize{ReLU/Sigm.} &\scriptsize{4} &\scriptsize{-} &\scriptsize{Keras \cite{I-4}/TensorFlow \cite{I-10}} \\
\scriptsize{FgSegNet-S \cite{P1C5-21680}} &\scriptsize{Current Image} &\scriptsize{Foreground} &\scriptsize{VGG-16  \cite{P1C5-2150-2}} &\scriptsize{TCNN/FPM} &\scriptsize{ReLU/Sigm.} &\scriptsize{4} &\scriptsize{-} &\scriptsize{Keras \cite{I-4}/TensorFlow \cite{I-10}} \\
\scriptsize{FgSegNet-V2 \cite{P1C5-21681}} &\scriptsize{Current Image} &\scriptsize{Foreground} &\scriptsize{VGG-16  \cite{P1C5-2150-2}} &\scriptsize{TCNN/FPM} &\scriptsize{ReLU/Sigm.} &\scriptsize{4} &\scriptsize{-} &\scriptsize{Keras \cite{I-4}/TensorFlow \cite{I-10}}  \\
\scriptsize{} &\scriptsize{} &\scriptsize{} &\scriptsize{} &\scriptsize{Feat. Fusions} &\scriptsize{} &\scriptsize{} &\scriptsize{} &\scriptsize{} \\
\scriptsize{MCSS \cite{P1C5-2179}} &\scriptsize{Backg.} &\scriptsize{Foreground} &\scriptsize{ConvNets \cite{P1C5-2100}} &\scriptsize{-} &\scriptsize{ReLU/Sigm.} &\scriptsize{2} &\scriptsize{2} &\scriptsize{-} \\
\scriptsize{} &\scriptsize{Current Image} &\scriptsize{} &\scriptsize{} &\scriptsize{} &\scriptsize{} &\scriptsize{} &\scriptsize{} &\scriptsize{} \\
\scriptsize{Guided Multi-scale CNN \cite{P1C5-2177}} &\scriptsize{Current Image} &\scriptsize{Foreground} &\scriptsize{ConvNets \cite{P1C5-2100}} &\scriptsize{Guided Learning} &\scriptsize{ReLU/Sigm.} &\scriptsize{4} &\scriptsize{-} &\scriptsize{-} \\
\hline
\scriptsize{\textbf{Fully CNN}} &\scriptsize{} &\scriptsize{} &\scriptsize{} &\scriptsize{} &\scriptsize{} &\scriptsize{} &\scriptsize{} &\scriptsize{} \\
\scriptsize{Fully CNNs \cite{P1C5-2160}} &\scriptsize{Backg. (Median)} &\scriptsize{Foreground} &\scriptsize{LeNet-5 \cite{P1C5-2100-1}} &\scriptsize{-} &\scriptsize{ReLU/Sigm.} &\scriptsize{4} &\scriptsize{-} &\scriptsize{Torch7} \\
\scriptsize{} &\scriptsize{Current Image} &\scriptsize{} &\scriptsize{} &\scriptsize{} &\scriptsize{} &\scriptsize{} &\scriptsize{} &\scriptsize{} \\
\scriptsize{Fully CNNs \cite{P1C5-2160}} &\scriptsize{Backg. (Median)} &\scriptsize{Foreground} &\scriptsize{ResNet \cite{P1C5-2166-2}} &\scriptsize{-} &\scriptsize{ReLU/Sigm.} &\scriptsize{-} &\scriptsize{-} &\scriptsize{Torch7} \\
\scriptsize{} &\scriptsize{Current Image} &\scriptsize{} &\scriptsize{} &\scriptsize{} &\scriptsize{} &\scriptsize{} &\scriptsize{} &\scriptsize{} \\
\scriptsize{Deep FCNNs \cite{P1C5-2165}} &\scriptsize{Current Image} &\scriptsize{Foreground} &\scriptsize{Multi. Branches (4)} &\scriptsize{CRF} &\scriptsize{PReLU \cite{P1C5-2165-3}} &\scriptsize{5 (Atrous)} &\scriptsize{1} &\scriptsize{-} \\
\scriptsize{MV-FCN \cite{P1C5-2170}} &\scriptsize{Current Image} &\scriptsize{Foreground} &\scriptsize{U-Net \cite{P1C5-2164-1}} &\scriptsize{2CFFs/PFF} &\scriptsize{ReLU/Sigm.} &\scriptsize{(2D Conv.)} &\scriptsize{1} &\scriptsize{Keras/Python} \\
\scriptsize{MFCN \cite{P1C5-2174}} &\scriptsize{Current Image} &\scriptsize{Foreground} &\scriptsize{VGG-16 \cite{P1C5-2150-2}} &\scriptsize{} &\scriptsize{ReLU/Sigm.} &\scriptsize{5} &\scriptsize{-} &\scriptsize{TensorFlow \cite{I-10}} \\
\scriptsize{CNN-SFC \cite{P1C5-2175}} &\scriptsize{3 For. Masks} &\scriptsize{Foreground} &\scriptsize{VGG-16 \cite{P1C5-2150-2}} &\scriptsize{} &\scriptsize{ReLU/Sigm.} &\scriptsize{13} &\scriptsize{None} &\scriptsize{TensorFlow \cite{I-10}} \\
\scriptsize{FCSN  \cite{P1C5-2178}} &\scriptsize{Backg. (SuBSENSE)} &\scriptsize{Foreground} &\scriptsize{FCN/VGG-16 \cite{P1C5-2165-1}} &\scriptsize{} &\scriptsize{ReLU/Sigm.} &\scriptsize{20} &\scriptsize{3} &\scriptsize{TensorFlow \cite{I-10}} \\
\scriptsize{} &\scriptsize{Current Image} &\scriptsize{} &\scriptsize{} &\scriptsize{} &\scriptsize{} &\scriptsize{} &\scriptsize{} &\scriptsize{} \\
\hline
\scriptsize{\textbf{Deep CNNs}} &\scriptsize{} &\scriptsize{} &\scriptsize{} &\scriptsize{} &\scriptsize{} &\scriptsize{} &\scriptsize{} &\scriptsize{} \\
\scriptsize{Deep CNN  \cite{P1C5-2140}} &\scriptsize{Backg. (SuBSENSE} &\scriptsize{Foreground} &\scriptsize{CNN} &\scriptsize{Multi-Layer} &\scriptsize{ReLU/Sigm.} &\scriptsize{3} &\scriptsize{-} &\scriptsize{-} \\
\scriptsize{} &\scriptsize{/FTSG)} &\scriptsize{} &\scriptsize{} &\scriptsize{Perceptron} &\scriptsize{} &\scriptsize{} &\scriptsize{} &\scriptsize{} \\
\scriptsize{} &\scriptsize{Current Image} &\scriptsize{} &\scriptsize{} &\scriptsize{(MLP)} &\scriptsize{} &\scriptsize{} &\scriptsize{} &\scriptsize{} \\
\scriptsize{TCNN/Joint TCNN \cite{P1C5-2161}} &\scriptsize{Backg.} &\scriptsize{Foreground} &\scriptsize{MCFC} &\scriptsize{DCGAN \cite{P1C5-2130-2}/} &\scriptsize{ReLU/Sigm.} &\scriptsize{-} &\scriptsize{-} &\scriptsize{Caffe \cite{I-1}/DeepLab \cite{P1C5-2130-4}} \\
\scriptsize{} &\scriptsize{Current Image} &\scriptsize{} &\scriptsize{(VGG-16)} &\scriptsize{Context Enc. \cite{P1C5-2130-1}} &\scriptsize{} &\scriptsize{} &\scriptsize{} &\scriptsize{} \\
\scriptsize{ADCNN \cite{P1C5-2162}} &\scriptsize{Current Image} &\scriptsize{Foreground} &\scriptsize{T-CNN} &\scriptsize{-} &\scriptsize{ReLU/Sigm.} &\scriptsize{7} &\scriptsize{None} &\scriptsize{Caffe \cite{I-1}} \\
\scriptsize{} &\scriptsize{} &\scriptsize{(Bound. Box)} &\scriptsize{S-CNN, C-CNN} &\scriptsize{} &\scriptsize{} &\scriptsize{} &\scriptsize{} &\scriptsize{} \\
\scriptsize{SFEN \cite{P1C5-2166}} &\scriptsize{Current Image} &\scriptsize{Foreground} &\scriptsize{VGG-16} &\scriptsize{Attention} &\scriptsize{ReLU/Sigm.} &\scriptsize{-} &\scriptsize{-} &\scriptsize{-} \\
\scriptsize{} &\scriptsize{} &\scriptsize{} &\scriptsize{GoogLeNet \cite{P1C5-2170-1}} &\scriptsize{ConvLSTM/} &\scriptsize{} &\scriptsize{} &\scriptsize{} &\scriptsize{} \\
\scriptsize{} &\scriptsize{} &\scriptsize{} &\scriptsize{ResNet} &\scriptsize{STN/CRF} &\scriptsize{} &\scriptsize{} &\scriptsize{} &\scriptsize{} \\
\hline
\scriptsize{\textbf{Structured CNN}} &\scriptsize{} &\scriptsize{} &\scriptsize{} &\scriptsize{} &\scriptsize{} &\scriptsize{} &\scriptsize{} &\scriptsize{} \\
\scriptsize{Struct CNN \cite{P1C5-2150}} &\scriptsize{Back. (Median)} &\scriptsize{Foreground} &\scriptsize{VGG-16} &\scriptsize{-} &\scriptsize{PReLU \cite{P1C5-2165-3}} &\scriptsize{13} &\scriptsize{-} &\scriptsize{Caffe \cite{I-1}} \\
\scriptsize{} &\scriptsize{Current Image t} &\scriptsize{} &\scriptsize{} &\scriptsize{} &\scriptsize{} &\scriptsize{} &\scriptsize{} &\scriptsize{} \\
\scriptsize{} &\scriptsize{Image t-1} &\scriptsize{} &\scriptsize{} &\scriptsize{} &\scriptsize{} &\scriptsize{} &\scriptsize{} &\scriptsize{} \\
\hline
\scriptsize{\textbf{3D CNNs}} &\scriptsize{} &\scriptsize{} &\scriptsize{} &\scriptsize{} &\scriptsize{} &\scriptsize{} &\scriptsize{} &\scriptsize{} \\
\scriptsize{3D ConvNet \cite{P1C5-2167}} &\scriptsize{10 Frames} &\scriptsize{Foreground} &\scriptsize{C3D Branch \cite{P1C5-2167-1}} &\scriptsize{-} &\scriptsize{-} &\scriptsize{6 (3D Conv.)} &\scriptsize{-} &\scriptsize{Caffe \cite{I-1}} \\
\scriptsize{STA-3D ConvNets (ReMoteNet) \cite{P1C5-2169}} &\scriptsize{Current Image} &\scriptsize{Foreground} &\scriptsize{Modified C3D} &\scriptsize{ST Attention} &\scriptsize{ReLU} &\scriptsize{(3D Conv.)} &\scriptsize{-} &\scriptsize{TensorFlow \cite{I-10}} \\
\scriptsize{} &\scriptsize{} &\scriptsize{(Bound. Box)} &\scriptsize{Branch \cite{P1C5-2169}} &\scriptsize{ConvLSTM} &\scriptsize{} &\scriptsize{} &\scriptsize{} &\scriptsize{} \\
\scriptsize{3D Atrous CNN \cite{P1C5-2170}} &\scriptsize{Current Image} &\scriptsize{Foreground} &\scriptsize{3D Atrous} &\scriptsize{-} &\scriptsize{ReLU} &\scriptsize{5 (3D Conv.)} &\scriptsize{-} &\scriptsize{TensorFlow \cite{I-10}} \\
\scriptsize{} &\scriptsize{} &\scriptsize{} &\scriptsize{ConvLSTM} &\scriptsize{} &\scriptsize{} &\scriptsize{} &\scriptsize{} &\scriptsize{} \\
\hline
\scriptsize{\textbf{Generative Adversarial Networks}} &\scriptsize{} &\scriptsize{} &\scriptsize{} &\scriptsize{} &\scriptsize{} &\scriptsize{} &\scriptsize{} &\scriptsize{} \\
\scriptsize{BScGAN \cite{P1C5-2191}} &\scriptsize{Back. (Median)} &\scriptsize{Foreground} &\scriptsize{cGAN \cite{P1C5-2191-1}} &\scriptsize{-} &\scriptsize{Leaky ReLU/Tanh} &\scriptsize{8} &\scriptsize{-} &\scriptsize{Pytorch} \\
\scriptsize{} &\scriptsize{Current Image} &\scriptsize{} &\scriptsize{Discrim. net} &\scriptsize{} &\scriptsize{Leaky ReLU/Sigm} &\scriptsize{4} &\scriptsize{-} &\scriptsize{Pytorch}  \\
\scriptsize{BGAN \cite{P1C5-2192}} &\scriptsize{Back. (Median)} &\scriptsize{Foreground} &\scriptsize{Bayesian GAN} &\scriptsize{-} &\scriptsize{-} &\scriptsize{-} &\scriptsize{-} &\scriptsize{-} \\
\scriptsize{} &\scriptsize{Current Image} &\scriptsize{} &\scriptsize{} &\scriptsize{} &\scriptsize{} &\scriptsize{} &\scriptsize{} &\scriptsize{} \\
\scriptsize{BPVGAN \cite{P1C5-2192}} &\scriptsize{Back. (Median)} &\scriptsize{Foreground} &\scriptsize{Paralell} &\scriptsize{-} &\scriptsize{-} &\scriptsize{-} &\scriptsize{-} &\scriptsize{-} \\
\scriptsize{} &\scriptsize{Current Image} &\scriptsize{} &\scriptsize{Bayesian GAN} &\scriptsize{} &\scriptsize{} &\scriptsize{} &\scriptsize{} &\scriptsize{} \\
\scriptsize{NUMOD \cite{P1C5-2194}} &\scriptsize{Current Image} &\scriptsize{Back.} &\scriptsize{GFCN} &\scriptsize{-} &\scriptsize{ReLU/Sigm.} &\scriptsize{-} &\scriptsize{-} &\scriptsize{-} \\
\scriptsize{} &\scriptsize{Illum. Image} &\scriptsize{} &\scriptsize{Bayesian GAN} &\scriptsize{} &\scriptsize{} &\scriptsize{} &\scriptsize{} &\scriptsize{}  \\
\scriptsize{} &\scriptsize{Foreground} &\scriptsize{} &\scriptsize{Bayesian GAN} &\scriptsize{} &\scriptsize{} &\scriptsize{} &\scriptsize{} &\scriptsize{} \\
\hline
\end{tabular}}
\caption{Deep Neural Networks Architecture in Background Subtraction: A Comparative Overview. "-" stands for "not indicated" by the authors.} \centering
\label{BGS-Overview2}
\end{table}

\begin{landscape} 
\begin{table}
\scalebox{0.65}{
\begin{tabular}{|l|l|l|l|l|l|l|l|l|l|} 
\hline
\scriptsize{Methods} &\scriptsize{Muti-scale} &\scriptsize{Training} &\scriptsize{Training} &\scriptsize{Spatial} &\scriptsize{Computation} &\scriptsize{End-to-End} &\scriptsize{Long-Term (Temporal)} &\scriptsize{Features} &\scriptsize{Type}\\
\scriptsize{} &\scriptsize{(Size)} &\scriptsize{(Over-fitting)} &\scriptsize{(GT)} &\scriptsize{(Pixel)} &\scriptsize{} &\scriptsize{} &\scriptsize{} &\scriptsize{}  &\scriptsize{}\\
\hline
\hline
\scriptsize{\textbf{Basic CNNs}} &\scriptsize{} &\scriptsize{} &\scriptsize{} &\scriptsize{} &\scriptsize{} &\scriptsize{} &\scriptsize{} &\scriptsize{} &\scriptsize{} \\
\scriptsize{ConvNets \cite{P1C5-2100}} &\scriptsize{No ($27 \times 27$)} &\scriptsize{Scene-specific} &\scriptsize{GT/IUTIS} &\scriptsize{No} &\scriptsize{Yes} &\scriptsize{No (Pre-proc.)} &\scriptsize{No} &\scriptsize{Grey} &\scriptsize{Generator}\\
\scriptsize{Basic CNNs \cite{P1C5-2180}} &\scriptsize{No ($64 \times 64$)} &\scriptsize{Scene-specific} &\scriptsize{GT} &\scriptsize{No} &\scriptsize{No} &\scriptsize{No (Pre-proc.)} &\scriptsize{No} &\scriptsize{RGB/IR} &\scriptsize{Generator}\\
\scriptsize{Basic CNNs \cite{P1C5-2162}} &\scriptsize{No ($31 \times 31$)} &\scriptsize{Scene-specific} &\scriptsize{GT} &\scriptsize{No} &\scriptsize{Yes} &\scriptsize{Yes} &\scriptsize{No} &\scriptsize{RGB} &\scriptsize{Generator} \\
\scriptsize{Basic CNNs \cite{P1C5-2182}} &\scriptsize{Frame} &\scriptsize{-} &\scriptsize{one GT} &\scriptsize{No} &\scriptsize{-} &\scriptsize{No (RPoTP)} &\scriptsize{Yes} &\scriptsize{RPoTP feature \cite{P1C5-2182}} &\scriptsize{Generator} \\
\scriptsize{Basic CNNs \cite{P1C5-2183}} &\scriptsize{Patch} &\scriptsize{-} &\scriptsize{GT (SBM-RGBD)} &\scriptsize{No} &\scriptsize{No} &\scriptsize{No (Pre-process.)} &\scriptsize{Np} &\scriptsize{Depth feature} &\scriptsize{Generator} \\
\hline
\scriptsize{\textbf{Multi-scale and Cascaded CNNs}} &\scriptsize{} &\scriptsize{} &\scriptsize{} &\scriptsize{} &\scriptsize{} &\scriptsize{} &\scriptsize{} &\scriptsize{} &\scriptsize{} \\
\scriptsize{Multi-scale CNNs \cite{P1C5-2162}} &\scriptsize{3 scales} &\scriptsize{Scene-specific} &\scriptsize{GT}  &\scriptsize{Cascaded (2)} &\scriptsize{Yes} &\scriptsize{Yes} &\scriptsize{No} &\scriptsize{RGB} &\scriptsize{Generator} \\
\scriptsize{Cascaded CNNs \cite{P1C5-2162}} &\scriptsize{3 scales}  &\scriptsize{Scene-specific} &\scriptsize{GT}  &\scriptsize{Cascaded (2 levels)} &\scriptsize{Yes} &\scriptsize{Yes} &\scriptsize{No} &\scriptsize{RGB} &\scriptsize{Generator} \\
\scriptsize{FgSegNet-M \cite{P1C5-2168}} &\scriptsize{3 scales}  &\scriptsize{Imbalanced data} &\scriptsize{GT}  &\scriptsize{TNN} &\scriptsize{18 fr/s} &\scriptsize{Yes} &\scriptsize{No} &\scriptsize{RGB} &\scriptsize{Generator} \\
\scriptsize{FgSegNet-S \cite{P1C5-21680}} &\scriptsize{FPM}  &\scriptsize{Imbalanced data} &\scriptsize{GT}  &\scriptsize{TNN} &\scriptsize{-} &\scriptsize{Yes} &\scriptsize{No} &\scriptsize{RGB} &\scriptsize{Generator} \\
\scriptsize{FgSegNet-V2 \cite{P1C5-21681}} &\scriptsize{M-FPM}  &\scriptsize{Imbalanced data} &\scriptsize{GT}  &\scriptsize{TNN} &\scriptsize{-} &\scriptsize{Yes} &\scriptsize{No} &\scriptsize{RGB} &\scriptsize{Generator} \\
\scriptsize{MCSS \cite{P1C5-2179}} &\scriptsize{3 scales ($27 \times 27$)}  &\scriptsize{Scene-specific} &\scriptsize{GT (Small Number)}  &\scriptsize{Cascaded (2 levels)} &\scriptsize{-} &\scriptsize{Yes} &\scriptsize{No} &\scriptsize{Grey} &\scriptsize{Generator} \\
\scriptsize{Guided Multi-scale \cite{P1C5-2177}} &\scriptsize{3 scales ($31 \times 31$)}  &\scriptsize{Scene-specific} &\scriptsize{GT}  &\scriptsize{-} &\scriptsize{-} &\scriptsize{No (Post-proc.)} &\scriptsize{No} &\scriptsize{RGB} &\scriptsize{Generator} \\
\hline
\scriptsize{\textbf{Fully CNNs}} &\scriptsize{} &\scriptsize{} &\scriptsize{} &\scriptsize{} &\scriptsize{} &\scriptsize{} &\scriptsize{} &\scriptsize{} &\scriptsize{}  \\
\scriptsize{Fully CNNs \cite{P1C5-2160}} &\scriptsize{No}  &\scriptsize{Scene-specific} &\scriptsize{GT}  &\scriptsize{No} &\scriptsize{Yes} &\scriptsize{Yes} &\scriptsize{No} &\scriptsize{Grey} &\scriptsize{Generator} \\
\scriptsize{Deep FCNNs \cite{P1C5-2165}} &\scriptsize{No}  &\scriptsize{-} &\scriptsize{GT}  &\scriptsize{Atrous} &\scriptsize{Yes} &\scriptsize{Yes} &\scriptsize{No} &\scriptsize{-(RGB?)} &\scriptsize{Generator} \\
\scriptsize{MV-FCN \cite{P1C5-2170}} &\scriptsize{Inception Mod.}  &\scriptsize{-} &\scriptsize{GT}  &\scriptsize{-} &\scriptsize{-} &\scriptsize{Yes} &\scriptsize{Encoder} &\scriptsize{-(RGB?)} &\scriptsize{Generator} \\
\scriptsize{MFCN \cite{P1C5-2174}} &\scriptsize{Yes ($224 \times 244 \times 3$)}  &\scriptsize{-} &\scriptsize{Mean}  &\scriptsize{-} &\scriptsize{27 fr/s} &\scriptsize{Yes} &\scriptsize{No} &\scriptsize{Infrared} &\scriptsize{Generator} \\
\scriptsize{} &\scriptsize{Yes ($224 \times244 \times 3$)}  &\scriptsize{-} &\scriptsize{Mean}  &\scriptsize{-} &\scriptsize{-} &\scriptsize{Yes} &\scriptsize{No} &\scriptsize{RGB} &\scriptsize{Generator} \\
\scriptsize{CNN-SFC \cite{P1C5-2175}} &\scriptsize{Semantic}  &\scriptsize{No} &\scriptsize{GT}  &\scriptsize{No} &\scriptsize{-} &\scriptsize{No} &\scriptsize{No} &\scriptsize{Black/White} &\scriptsize{Generator} \\
\scriptsize{FCSN  \cite{P1C5-2178}} &\scriptsize{Semantic}  &\scriptsize{No} &\scriptsize{GT/SuBSENSE}  &\scriptsize{Semantic} &\scriptsize{48 fr/s} &\scriptsize{Yes} &\scriptsize{No} &\scriptsize{-(RGB?)} &\scriptsize{Generator} \\
\hline
\scriptsize{\textbf{Deep CNNs}} &\scriptsize{} &\scriptsize{} &\scriptsize{} &\scriptsize{} &\scriptsize{} &\scriptsize{} &\scriptsize{} &\scriptsize{}  &\scriptsize{}  \\
\scriptsize{Deep CNN  \cite{P1C5-2140}} &\scriptsize{No ($37 \times 37$)}  &\scriptsize{Scene-specific} &\scriptsize{GT}  &\scriptsize{No} &\scriptsize{Yes} &\scriptsize{No (Post-proc.)} &\scriptsize{No} &\scriptsize{RGB)} &\scriptsize{Generator} \\
\scriptsize{TCNN/Joint TCNN \cite{P1C5-2161}} &\scriptsize{Yes ($961 \times 961$)}  &\scriptsize{Background} &\scriptsize{GT}  &\scriptsize{No} &\scriptsize{5 fr/s} &\scriptsize{Yes} &\scriptsize{No} &\scriptsize{RGB} &\scriptsize{Generator} \\
\scriptsize{} &\scriptsize{Atrous Sampling Rate}  &\scriptsize{Generation} &\scriptsize{(PASCAL VOC 2012)}  &\scriptsize{} &\scriptsize{} &\scriptsize{} &\scriptsize{} &\scriptsize{} &\scriptsize{} \\
\scriptsize{ADCNN \cite{P1C5-2162}} &\scriptsize{Yes}  &\scriptsize{Discriminative} &\scriptsize{GT}  &\scriptsize{No} &\scriptsize{-}&\scriptsize{Yes} &\scriptsize{No} &\scriptsize{RGB} &\scriptsize{Generator} \\
\scriptsize{} &\scriptsize{}  &\scriptsize{Features} &\scriptsize{(CUHK, MIT, PETS)}  &\scriptsize{} &\scriptsize{} &\scriptsize{} &\scriptsize{} &\scriptsize{} &\scriptsize{} \\
\scriptsize{SFEN \cite{P1C5-2166}} &\scriptsize{Semantic}  &\scriptsize{No} &\scriptsize{GT}  &\scriptsize{STN} &\scriptsize{15 fr/s} &\scriptsize{Yes} &\scriptsize{No} &\scriptsize{RGB} &\scriptsize{Generator} \\
\scriptsize{SFEN+CRF \cite{P1C5-2166}} &\scriptsize{Semantic}  &\scriptsize{No} &\scriptsize{GT}  &\scriptsize{STN/CRF} &\scriptsize{6 fr/s} &\scriptsize{Yes} &\scriptsize{No} &\scriptsize{RGB} &\scriptsize{Generator} \\
\scriptsize{SFEN+PSL+CRF \cite{P1C5-2166}} &\scriptsize{Semantic ($224 /times 224$)}  &\scriptsize{No} &\scriptsize{GT}  &\scriptsize{STN/CRF/PSL} &\scriptsize{5 fr/s} &\scriptsize{Yes} &\scriptsize{ConvLSTM} &\scriptsize{RGB} &\scriptsize{Generator} \\
\hline
\scriptsize{\textbf{Structured CNNs}} &\scriptsize{} &\scriptsize{} &\scriptsize{} &\scriptsize{} &\scriptsize{} &\scriptsize{} &\scriptsize{} &\scriptsize{}  &\scriptsize{} \\
\scriptsize{Struct CNN \cite{P1C5-2150}} &\scriptsize{Contours ($336 \times 336$)}  &\scriptsize{No} &\scriptsize{GT}  &\scriptsize{Superpixel} &\scriptsize{-} &\scriptsize{No (Post-proc)} &\scriptsize{No} &\scriptsize{Grey} &\scriptsize{Generator} \\
\scriptsize{\textbf{3D CNNs}} &\scriptsize{} &\scriptsize{} &\scriptsize{} &\scriptsize{} &\scriptsize{} &\scriptsize{} &\scriptsize{} &\scriptsize{} &\scriptsize{} \\
\scriptsize{3D ConvNet \cite{P1C5-2167}} &\scriptsize{Multi-kernel upsampling}  &\scriptsize{Yes} &\scriptsize{GT}  &\scriptsize{No}$$&\scriptsize{-} &\scriptsize{Yes} &\scriptsize{3D} &\scriptsize{-(RGB?)} &\scriptsize{Generator} \\
\scriptsize{STA-3D ConvNets (ReMoteNet) \cite{P1C5-2169}} &\scriptsize{$1280 \times 720$)}  &\scriptsize{No} &\scriptsize{GT}  &\scriptsize{STA ConvLSTM} &\scriptsize{Fast} &\scriptsize{Yes} &\scriptsize{STA ConvLSTM} &\scriptsize{RGB} &\scriptsize{Generator} \\
\scriptsize{3D Atrous CNN \cite{P1C5-2170}} &\scriptsize{$320 \times 240$)}  &\scriptsize{No} &\scriptsize{GT}  &\scriptsize{Atrous} &\scriptsize{-} &\scriptsize{Yes} &\scriptsize{3D/ConvLSTM} &\scriptsize{-(RGB?)} &\scriptsize{Generator} \\
\hline
\scriptsize{\textbf{Generative Adversarial Networks}} &\scriptsize{} &\scriptsize{} &\scriptsize{} &\scriptsize{} &\scriptsize{} &\scriptsize{} &\scriptsize{} &\scriptsize{} &\scriptsize{} \\
\scriptsize{BScGAN \cite{P1C5-2191}}  &\scriptsize{$256 \times 256$)}  &\scriptsize{No} &\scriptsize{GT}  &\scriptsize{No} &\scriptsize{10 fr/s} &\scriptsize{Yes} &\scriptsize{No} &\scriptsize{-(RGB?)} &\scriptsize{Generator/Discriminator} \\
\scriptsize{BGAN \cite{P1C5-2192}} &\scriptsize{-}  &\scriptsize{-} &\scriptsize{GT}  &\scriptsize{-} &\scriptsize{-} &\scriptsize{Yes} &\scriptsize{-} &\scriptsize{-} &\scriptsize{Generator/Discriminator} \\
\scriptsize{BPVGAN \cite{P1C5-2192}} &\scriptsize{-}  &\scriptsize{-} &\scriptsize{GT}  &\scriptsize{-} &\scriptsize{Parallel Implem.} &\scriptsize{Yes} &\scriptsize{-} &\scriptsize{-} &\scriptsize{Generator/Discriminator} \\
\scriptsize{NUMOD \cite{P1C5-2194}} &\scriptsize{Frame}  &\scriptsize{No} &\scriptsize{$I=B+C+F$}  &\scriptsize{No} &\scriptsize{-} &\scriptsize{Yes} &\scriptsize{No} &\scriptsize{RGB} &\scriptsize{Generator/} \\
\hline
\end{tabular}}
\caption{Deep Neural Networks in Background Subtraction: A Comparative Overview for Challenges. "-" stands for "not indicated" by the authors.} \centering
\label{BGS-Overview3}
\end{table}
\end{landscape} 

\section{Background Subtraction}
\label{BackgroundSubtraction}
Background subtraction consists of comparing the background image with the current image to label pixels as background or foreground pixels. The top algorithms on the large-scale dataset CDnet 2014 are three DNNs based methods (FgSegNet \cite{P1C5-2150}, BSGAN \cite{P1C5-2172}, Cascaded CNN \cite{P1C5-2110}) for supervised approaches followed by three no-supervised methods that are multi-features/multi-cues approaches (SuBSENSE \cite{P2C0-40}, PAWCS \cite{P2C0-50}, IUTIS \cite{P1C5-2300}). This task is a classification one, that can be achieved with success by DNN. For this, different methods have been developed in literature and we review them in the following sub-sections. Table \ref{BGS-Overview1} shows an overview of these methods. In addition, the list of publications is available at the Background Subtraction Website\protect\footnotemark[14] and is regularly updated.

\footnotetext[14]{{https://sites.google.com/site/backgroundsubtraction/recent-background-modeling/deep-learning}}

\subsection{Convolutional Neural Networks}
Braham and Van Droogenbroeck \cite{P1C5-2100} were the first authors to use Convolutional Neural Networks (CNNs) for background subtraction. This model named ConvNet has a similar structure than LeNet-5 \cite{P1C5-2100-1}. Thus, the  background subtraction model involves four stages: background image extraction via a temporal median in grey scale, specific-scene dataset generation, network training and background subtraction. More precisely, the background model is built for a specific scene. For each frame in a video sequence, image patches that are centered on each pixel are extracted and then they are combined with corresponding patches from the background model. Braham and Van Droogenbroeck \cite{P1C5-2100} used a patch size of $27 \times 27$. After, these combined patches are fed to the network to predict probability of foreground pixels. For the architecture, Braham and  Van Droogenbroeck \cite{P1C5-2100} employed  $5 \times 5$ local receptive fields, and $3 \times 3$ non-overlapping receptive fields for all pooling layers. The first and second convolutional layers have $6$ and $16$ feature maps, respectively. The first fully connected layer has $120$ hidden units and the output layer consists of a single sigmoid unit. The algorithm needs for training the foreground results of a previous segmentation algorithm named IUTIS \cite{P1C5-2300} or the ground truth information provided in CDnet 2014 \cite{Dataset-1}.  Half of the training examples are used for training ConvNet and the remaining frames are used for testing. By using the results of the IUTIS method \cite{P1C5-2300}, the segmentation produced by the ConvNet is very similar to other state-of-the-art methods whilst the algorithm outperforms all other methods significantly when using the ground-truth information especially in videos of hard shadows and night videos. With the CDnet2014 dataset (excluding the IOM and PTZ categories), this method with IUTIS and GT achieved an average F-Measure of $0.7897$ and $0.9046$, respectively. Baustita et al. \cite{P1C5-2120} also used a simple CNN but for the specific task of vehicle detection. For pedestrian detection, Yan et al. \cite{P1C5-2180} employed the similar scheme with both visible and thermal images. Then, the inputs of the network have a size of $64  \times 64 \times 8$ which includes the visible frame (RGB), thermal frame (IR), visible background (RGB) and thermal background (IR). The outputs of the network have a size of $64 \times 64 \times 2$. Experiments on OCTBVS dataset\protect\footnotemark[15] show that this method outperforms T2-FMOG \cite{7700}, SuBSENSE \cite{P2C0-40}, and DECOLOR \cite{RPCA-25}. For biodiversity detection in terrestrial and marine environments, Weinstein  \cite{P1C5-2181} employed the GoogLeNet architecture integrated in a software called DeepMeerkat\protect\footnotemark[16]. Experiments on humming bird videos show robust performance in challenging outdoor scenes where moving foliages occur.

\footnotetext[15]{{http://vcipl-okstate.org/pbvs/bench/}}
\footnotetext[16]{{http://benweinstein.weebly.com/deepmeerkat.html}}

\textbf{Remarks:} ConvNet is the simplest manner to learn the differences between the background and the foreground via CNNs. Thus, the work of Braham and Van Droogenbroeck \cite{P1C5-2100} presents the very big merit to be the first application of deep learning for background subtraction, and can then be used as a reference for comparison in terms of improvements and performance. But, it presents several limitations: 1) It is difficult to learn the high-level information through patches \cite{P1C5-2178}; 2) due to the over-fitting that is caused by using highly redundant data for training, the network is scene-specific. In practice, it can only process a certain scenery, and needs to be retrained for other video scenes \cite{P1C5-2140}. This fact is not a problem most of the time because the camera is fixed filming always similar scenes. But, it may not be the case in certain applications as pointed out by Hu et al. \cite{P1C5-2176}. ; 3) Each pixel is processed independently and then the foreground mask may contain isolated false positives and false negatives; 4) It is computationally expensive due to large number of patches extracted from each frame as remarked by Lim and Keles \cite{P1C5-2168}; 5) it requires pre-processing or post-processing of the data, and hence is not based on an end-to-end learning framework \cite{P1C5-2176}; 6) ConvNet use few frames as input and thus can not consider long-term dependencies of the input video sequences \cite{P1C5-2176}; and 7) ConvNet is  a deep encoder-decoder network that is a generator network. But, the classical generator networks produce blurry foreground regions and such networks can not preserve the objects edges because they minimize the classical loss functions (e.g., Euclidean distance) between the predicted output and the ground-truth \cite{P1C5-2178}. Since this first valuable work, the posterior methods developed in the literature attempt to alleviate these limitations that are the main challenges to use DNN in background subtraction. Table \ref{BGS-Overview2} shows a comparative overview with all the posterior methods while Table \ref{BGS-Overview3} show an overview in terms of the challenges. These tables are discussed in Section \ref{Discussion}.\\

\subsection{Multi-scale and Cascaded CNNs}
Wang et al. \cite{P1C5-2110} proposed a deep learning method for an iterative ground-truth generation process in the context of background modeling algorithms validation. In order to yield the ground truths, this method segments the foreground objects by learning the appearance of foreground samples. First, Wang et al. \cite{P1C5-2110}  designed basic CNN and the multi-scale CNN which processed each pixel independently based on the information contained in their local patch of size 31*31 in each channel RGB. The basic CNN model consists of 4 convolutional layers and 2 fully connected layers. The first 2 convolutional layers come with 2*2 max pooling layer. Each convolutional layer uses a filter size of $7 \times 7$ and Rectified Linear Unit (ReLU) as the activation function. By considering the CNN output as a likelihood probability, a cross entropy loss function is employed for training. Because, this basic model processes patches of size $31 \times 31$, its performance is limited to distinguish foreground and background objects with the same size or less. This limitation is alleviated by the multi-scale CNN model which gives three outputs of three different sizes further combined in the original size.  In order to model the dependencies among adjacent pixels and thus enforce spatial coherence, Wang et al. \cite{P1C5-2110} employed the multi-scale CNN model with a cascaded architecture that is named Cascaded CNN. Practically, the CNN presents the advantage of learning or extracting its own features that may be better than hand-designed features. The CNN is fed with manually generated foreground objects from some frames of a video sequence to learn the foreground features. After this step, the CNN employs generalization to segment the remaining frames of the video. Wang et al. \cite{P1C5-2110} trained scene specific networks using $200$ frames by manual selection. Cascaded CNN provides an overall F-Measure of $0.9209$ in CDnet2014 dataset \cite{Dataset-1}. For the Cascaded CNN's implementation\protect\footnotemark[17] available online, Wang et al. \cite{P1C5-2110} used the Caffe library\protect\footnotemark[18] \cite{I-1} and MatConvNet\protect\footnotemark[19]. The limitations of Cascaded CNN are as follows: 1) it is more dedicated to ground-truth generation than an automated background/foreground separation method, and 2) it is also computationally expensive. 

Lim and Keles \cite{P1C5-2168} proposed a method called FgSegNet-M\protect\footnotemark[20] based on a triplet CNN and a Transposed Convolutional Neural Network (TCNN) attached at the end of it in an encoder-decoder structure. Practically, the four blocks of the pre-trained VGG-16 \cite{P1C5-2150-2} Net is employed at the beginning of the proposed CNNs under a triplet framework as the multiscale feature encoder. Furthermore, a decoder network is integrated at the end of it to map the features to a pixel-level foreground probability map. Then, a threshold is applied to this map to obtain binary segmentation labels. Practically, Lim and Keles \cite{P1C5-2168} generated scene specific models using only a few frames (to 50 up to 200) similar to Wang et al. \cite{P1C5-2110}. Experimental results \cite{P1C5-2168} show that TCNN outperforms both ConvNet  \cite{P1C5-2100} and Cascaded CNN \cite{P1C5-2110}, and practically outperformed all the reported methods by an overall F-Measure of $0.9770$. In a further work, Lim and Keles \cite{P1C5-21680} designed a variant of FgSegNet-M called FgSegNet-S by adding a feature pooling module FPM which operates on top of the final encoder (CNN) layer. In an additional work, Lim et al. \cite{P1C5-21681} proposed a modified FM with feature fusion. This last version called FgSegNet-V2\protect\footnotemark[21] ranked as number one on the CDnet 2014 dataset.

These previous methods usually require a large amount of densely labeled video training data. To solve this problem, Liao et al. \cite{P1C5-2179} designed a multi-scale cascaded scene-specific (MCSS) CNNs based background subtraction method with a novel training strategy. The architecture combined the ConvNets \cite{P1C5-2100} and the multiscale-cascaded architecture \cite{P1C5-2110} with a training that takes advantage of the balance of positive and negative training samples. Experimental results show that MCSS outperforms Deep CNN \cite{P1C5-2140}, TCNN \cite{P1C5-2161} and SFEN \cite{P1C5-2166} with a score of $0.904$ on the CDnet 2014 dataset by excluding the PTZ category.

Liang et al. \cite{P1C5-2177} developed a multi-scale CNN based background subtraction method by learning a specific CNN model for each video to ensure accuracy, but manage to avoid manual labeling. First, Liang et al. \cite{P1C5-2177} applied the SubSENSE algorithm to get an initial foreground mask. Then, an adaptive strategy is applied to select reliable pixels to guide the CNN training because the outputs of SubSENSE cannot be directly used as ground truth due the lack of accuracy of the results. A simple strategy is also proposed to automatically select informative frames for the guided learning. Experiments on the CDnet 2014 dataset show that Guided Multi-scale CNN gives a better F-Measure of $0.7591$ than DeepBS \cite{P1C5-2140} and SuBSENSE \cite{P2C0-40}.

\footnotetext[17]{{https://github.com/zhimingluo/MovingObjectSegmentation/}}
\footnotetext[18]{{http://caffe.berkeleyvision.org/tutorial/solver.html}}
\footnotetext[19]{{http://www.vlfeat.org/matconvnet/}}
\footnotetext[20]{{https://github.com/lim-anggun/FgSegNet}}
\footnotetext[21]{{https://github.com/lim-anggun/FgSegNet-v2}}

\subsection{Fully CNNs}
Cinelli \cite{P1C5-2160} proposed a similar method than Braham and Droogenbroeck \cite{P1C5-2100} by exploring the advantages of Fully Convolutional Neural Networks (FCNNs) \cite{P1C5-2165-1} to diminish the computational requirements. FCNN use convolutional layer to replace the fully connected layer in traditional convolution networks, which can avoid the disadvantages caused by fully connection layer. Practically, Cinelli tested both LeNet5 \cite{P1C5-2100-1} and ResNet \cite{P1C5-2100-2} architectures.  As the ResNet presents a greater degree of hyper-parameter setting (namely the size of the model and even the organization of layers) compare to LeNet5, Cinelli also varied different features of the ResNet architectures to optimize them for background/foreground separation. For this, Cinelli used the networks designed for the ImageNet Large Scale Visual Recognition Challenge (ILSVRC \protect\footnotemark[22]), which deal with $224 \times 224$ pixel images, and  those for the CIFAR-10 and CIFAR-100 datasets\protect\footnotemark[23], which have $32 \times 32$ pixel-images as input.  The FAIR\protect\footnotemark[24] implementation is employed. From this study, the best models on the CDnet 2014 dataset  \cite{Dataset-1} are the $32$-layer CIFAR-derived dilated network and the pre-trained 34-layer ILSVRC-based dilated model adapted by direct substitution. But, Cinelli \cite{P1C5-2160} only provided visual results without F-measure. 

\footnotetext[22]{{http://www.image-net.org/challenges/LSVRC/}}
\footnotetext[23]{{https://www.cs.toronto.edu/~kriz/cifar.html}}
\footnotetext[24]{{https://github.com/facebook/fb.resnet.torch}}

In another work, Yang et al. \cite{P1C5-2165} also used FCNN but with a structure of shortcut connected block with multiple branches. Each block provides four different branches. Practically, the front of three branches calculate different features by using different atrous convolution, and the last branch is the shortcut connection. For the spatial information, atrous convolution \cite{P1C5-2165-2} is employed instead of common convolution in order to miss considerable details by expanding the receptive fields. For the activation layers, PReLU Parametric Rectified Linear Unit (PReLU) \cite{P1C5-2165-3} introduced a learned parameter to transform the values less than $0$. Yang et al. \cite{P1C5-2165} also employed a refinement method using Conditional Random Fields (CRF). Experimental results show that this method outperforms traditional background subtraction methods (MOG \cite{P1C2-MOG-10} and Codebook \cite{2}) as well as recent state-of-art methods (ViBe \cite{3}, PBAS \cite{P2C1-230} and P2M \cite{4}) on the CDnet 2012 dataset \cite{Dataset-2}. But, Yang et al. \cite{P1C5-2165} evaluated their method on a subset of $6$ sequences of CDnet 2012 \cite{Dataset-2} instead of all the categories of CDnet 2014 \cite{Dataset-1} making the comparison more difficult with the other DNN methods.

Alikan \cite{P1C5-2170} designed a Multi-View receptive field Fully CNN (MV-FCN) based on fully convolutional structure, inception modules \cite{P1C5-2166-1}, and residual networking. MV-FCN is based on inception module \cite{P1C5-2170-1} designed by Google that performs convolution of multiple filters with different scales on the same input to simulate human cognitive processes in perceiving
multi-scale information, and ResNet \cite{P1C5-2100-2} developed by Microsoft that acts as lost feature recovery mechanism. In addition, Alikan \cite{P1C5-2170} exploits intra-domain transfer learning that boosts the correct foreground region prediction. Practically, MV-FCN employs inception modules at early and late stages with three different sizes of receptive fields to capture invariance at various scales. The features learned in the encoding phase are fused with appropriate feature maps in the decoding phase through residual connections for achieving enhanced spatial representation. These multi-view receptive fields and residual feature connections provide generalized features for a more accurate pixel-wise foreground region identification. The training is made with the CDnet 2014 \cite{Dataset-1}. Alikan et al. \cite{P1C5-2170} evaluated MV-FCN against classical neural networks (Stacked Multi-Layer \cite{P1C5-420},  Multi-Layered SOM \cite{P1C5-421}), and two deep learning approaches (SDAE \cite{P1C5-2010}, Deep CNN \cite{P1C5-2140}) on the CDnet 2014 \cite{Dataset-1} but only on selected sequences making the comparison less complete. 

Zeng and Zhu \cite{P1C5-2174} developed a Multiscale Fully Convolutional Network (MFCN) for moving object detection in infrared videos. MFCN does not need to extract the background images. The input is frames from different sequences, and the output is
a probability map. Practically, Zeng and Zhu \cite{P1C5-2174} used the VGG-16 as architecture and  the inputs have a size of $224 \times 224$. The VGG-16 network is split into five blocks with each block containing some convolution and max pooling operations. The The lower blocks have a higher spatial resolution and contain more low-level local features whilst the deeper blocks contain more
high-level global features at a lower resolution. A contrast layer is added behind the output feature layer based on the average pooling operation with a kernel size of $3 \times 3$. In order to exploit multi-scale features from multiple layers, 
Zeng and Zhu \cite{P1C5-2174} employed  a set of deconvolution operations to up-sample the features, creating an output probability map the same size as the input. For the loss function, the cross-entropy is used. The layers from VGG-16 are initialized
with pre-trained weights, whilst the other weights are randomly initialized with a truncated normal distribution. The adam optimizer method is used for updating the model parameters. Experimental results on the THM category of CDnet 2014 \cite{Dataset-1} show that MFCN obtains the best score in this category with $0.9870$ while  Cascaded CNN \cite{P1C5-2110} obtains $0.8958$ whilst MFCN achieves a score of $0.96$ over all the categories. In a further work, Zeng and Zhu \cite{P1C5-2175} fused the results produced by different background subtraction algorithms (SuBSENSE \cite{P2C0-40}, FTSG \cite{P2C0-500}, and CwisarDH+ \cite{P1C5-1604}) in order to output a more precise result. This method called CNN-SFC outperforms its direct competitor IUTIS \cite{P1C5-2300} on the CDnet 2014 dataset.
 
Lin et al. \cite{P1C5-2178} designed a deep Fully Convolutional Semantic Network (FCSN) for background subtraction. First, FCN is able to learn the global differences between the foreground and the background. Second, SuBSENSE \cite{P2C0-40} algorithm is employed to generate robust background image with better performance, which is concatenated into the input of the network together with
the video frame. Furthermore, Lin et al. \cite{P1C5-2178} initialized the weights of FCSN by partially using pre-trained weights
of FCN-VGG16, because these weights are applied to semantic segmentation. Then, FCSN can understand semantic information of images and converge faster. In addition, FCSN uses less training data and get better result with the help of pre-trained weights. \\

\subsection{Deep CNNs}
Babaee et al. \cite{P1C5-2140} proposed a deep CNNs based moving objects detection method which contains the following components: an algorithm for background initialization via an average model in RGB, a CNN model for background subtraction, and a post-processing module of the networks output using a spatial median filter. First,  Babaee et al. \cite{P1C5-2140} proposed to distinguish the foreground pixels and background pixels with SuBSENSE algorithm \cite{P2C0-40}, and then only used the background pixel values to obtain the background average model. In order to have adaptive memory length based on the motion of the camera and objects in the video frames, Babaee et al. \cite{P1C5-2140} used Flux Tensor with Split Gaussian Models (FTSG \cite{P2C0-500}) algorithm. For the network architecture and training, Babaee et al. \cite{P1C5-2140} trained the CNNs with background images obtained by the SuBSENSE algorithm \cite{P2C0-40}. With images of size $240 \times 320$ pixels, the network is trained with pairs of RGB image patches (triplets of size $37 \times 37$) from video, background frames and the respective ground truth segmentation patches (CDnet 2014 \cite{Dataset-1} with around $5\%$ of the data). Thus, instead of training a network for a specific scene, Babaee et al. \cite{P1C5-2140} trained their model all at once by combining training frames from various video sequences including 5\% of frames from each video sequence. On the other hand, the same training procedure than ConvNet \cite{P1C5-2100} is employed. Each image-patches are combined with background-patches then fed to the network. The network contains $3$ convolutional layers and a $2$-layer Multi-Layer Perceptron (MLP). Rectified Linear Unit (ReLU) \cite{DNN-120} is used as activation function after each convolutional layer and the sigmoid function after the last fully connected layer. In addition, batch normalization layers  are used before each activation layer to decrease over-fitting and to also provide higher learning rates for training. Finally, a spatial-median filtering is applied in the post-processing step. This method provided foreground mask more precise than  ConvNet \cite{P1C5-2100} and not very prone to outliers in presence of dynamic backgrounds. Finally, deep CNN based background subtraction outperforms the existing algorithms when the challenge does not lie in the background modeling maintenance. Deep CNN obtained an F-Measure of $0.7548$ in CDnet2014 dataset \cite{Dataset-1}. The limitations of Deep CNN are as follows: 1) It can not well handle the camouflage regions within foreground objects, 2) it provided poor performance for PTZ videos, and 3) due to the corruption of the background images, it performs poorly in presence of large changes in the background.  

In a further work, Zhao et al. \cite{P1C5-2161} proposed an end-to-end two-stage deep CNN (TS-CNN) framework. In the first stage, a convolutional encoder-decoder sub-network is used to reconstruct the background images and encode rich prior knowledge of background scenes whilst the reconstructed background and current frame are the inputs into a Multi-Channel Fully-Convolutional sub-Network (MCFCN) for accurate foreground detection in the second stage. In the two-stage CNN, the reconstruction loss and segmentation loss are jointly optimized. Practically, the encoder contains a set of convolutions, and represents the input image as a latent feature vector. The decoder restores the background image from the feature vector. The $l_2$ loss was employed as the reconstruction loss. After training, the encoder-decoder network separates the background from the input image and restores a clean background image. The  second network can learn semantic knowledge of the foreground and background. Therefore, it could handle various challenges such as the night light, shadows and camouflaged foreground objects. Experimental results \cite{P1C5-2161} show that the TS-CNN outperforms SuBSENSE \cite{P2C0-40}, PAWCS \cite{P2C0-50}, FTSG \cite{P2C0-500} and SharedModel \cite{P2C0-600} in the case of night videos, camera jitter, shadows, thermal imagery and bad weather. In CDnet2014 dataset \cite{Dataset-1}, TS-CNN and Joint TS-CNN obtained an F-Measure of $0.7870$ and $0.8124$, respectively. 

In another approach, Li et al. \cite{P1C5-2162} designed an adaptive deep CNN (ADCNN) to predict object locations in a surveillance
scene. First, the generic CNN-based classifier is transfered to the surveillance scene by selecting useful kernels. Secondly, 
the context information of the surveillance scene is learned in the regression model for accurate location prediction. Our main contributions. ADCNN achieved very interesting performance on several surveillance datasets for pedestrian detection and vehicle detection but ADCNN focus on object detection and thus not use the principle of background subtraction. Furthermore, Li et al. \cite{P1C5-2162} provided results with the CUHK square dataset \cite{P1C5-2162-1}, the MIT traffic dataset \cite{P1C5-2162-2} and the PETS 2007\protect\footnotemark[25] instead of the CDnet2014 dataset \cite{Dataset-1}. 

\footnotetext[25]{{http://www.cvg.reading.ac.uk/pets2007/data.html}}

In another work, Chen et al. \cite{P1C5-2166} proposed to detect moving objects via an end-to-end deep sequence learning architecture with the pixel-level semantic features. Video sequences are the input into a deep convolutional encoder-decoder network to extract pixel-level semantic features. Practically,  Chen et al. \cite{P1C5-2166} used the VGG-16 \cite{P1C5-2150-2} as encoder-decoder network but other architectures, such as GoogLeNet \cite{P1C5-2166-1}, ResNet50 \cite{P1C5-2100-2} can be also used into this framework. An attention long short-term memory model named Attention ConvLSTM is used to integrate pixel-wise changes over time. After, a Spatial Transformer Network (STN) model and a Conditional Random Fields (CRF) layer are employed to reduce the sensitivity to camera motion and to smooth the foreground boundaries, respectively. Experimental results \cite{P1C5-2166} on the two large-scale dataset CDnet 2014 dataset \cite{Dataset-1} and LASIESTA \cite{Dataset-3} show that the proposed method obtained similar results than Convnet \cite{P1C5-2100} with better performance for the category "Night videos", "Camera jitter", "Shadow" and "Turbulence". Attention ConvLSTM obtained an F-Measure of $0.8292$ with VGG-16, $0.7360$ with GoogLeNet and $0.8772$ with ResNet50. \\

\subsection{Structured CNNs}
Lim et al. \cite{P1C5-2150} developed an encoder-encoder structured CNN (Struct-CNN) for background subtraction. Thus, the background subtraction model involves the following components: a background image extraction via a temporal median in RGB, network training, background subtraction and foreground extraction based on super-pixel information. The structure is similar to the VGG16 network \cite{P1C5-2150-2} after excluding the fully connected layers. The encoder converts the $3$ (RGB) channel input (images of size $336 \times 336$ pixels) into  $512$-channel feature vector through convolutional and max-pooling layers yielding a $21 \times 21 \times 512$ feature vector. Then, the decoder converts the feature vector into a $1$-channel image of size  $336 \times 336$ pixels providing the foreground mask through deconvolutional and unpooling layers. Lim et al.\cite{P1C5-2150} trained this encoder-decoder structured network in the end-to-end manner using CDnet 2014 \cite{Dataset-1}. For the architecture, the decoder consists of $6$ deconvolutional layers and $4$ unpooling layers. In all deconvolutional layers, except for the last one, features are batch-normalized and the Parametric Rectified Linear Unit (PReLU) \cite{P1C5-2165-2} is employed as an activation function. The last deconvolutional layer which is the prediction layer used the sigmoid activation function to normalize outputs and then to provide the foreground mask. $5 \times 5$ kernels are used in all convolutional while a $3 \times 3$ kernel is employed in the prediction layer. In order to suppress the incorrect boundaries and holes in the foreground mask, Lim et al. \cite{P1C5-2150} used the superpixel information obtained by an edge detector. Experimental results \cite{P1C5-2150} show that Struct-CNN outperforms SuBSENSE \cite{P2C0-40}, PAWCS \cite{P2C0-50}, FTSG \cite{P2C0-500} and SharedModel \cite{P2C0-600} in the case of bad weather, camera jitter, low frame rate, intermittent object motion and thermal imagery.  Struct-CNN obtained an F-Measure of $0.8645$ on the CDnet 2014 dataset \cite{Dataset-1} excluding the "PTZ" category. Lim et al. \cite{P1C5-2150} excluded this category arguying that they focused only on static cameras. \\

\subsection{3D-CNNs}
Sakkos et al. \cite{P1C5-2167} designed an end-to-end 3D-CNN to track temporal changes in video sequences avoiding the use of a background model for the training. 3D-CNN can handle multiple scenes without further fine-tuning on each scene individually. For the architecture, Sakkos et al. \cite{P1C5-2167} used C3D branch \cite{P1C5-2167-1}. Experimental results \cite{P1C5-2167} reveal that 3D-CNN provides better performance than ConvNet \cite{P1C5-2100} and deep CNN \cite{P1C5-2140}. Furthermore, experiments on the ESI dataset \cite{6} which present extreme and sudden illumination changes, show that 3D-CNN outperforms two designed illumination invariant background subtraction methods that are Universal Multimode Background Subtraction (UMBS) \cite{5} and ESI \cite{6}. 3D-CNN obtained an average F-Measure of $0.9507$ in CDnet 2014 dataset. 

Yu et al. \cite{P1C5-2169} employed a spatial-temporal attention-based 3D ConvNets to jointly model the appearance and motion of objects-of-interest in a video for a  Relevant Motion Event detection Network (ReMotENet). The architecture is based on the C3D branch \cite{P1C5-2167-1}. But, instead of using max pooling both spatially and temporally, Yu et al. \cite{P1C5-2169} separated the spatial and temporal max pooling in order to capture fine-grained temporal information, and makes the network deeper to learn better representations. Experiments demonstrate that ReMotENet achieves comparable or even better performance, but is three to four orders of magnitude faster than the object detection based method. It can detect relevant motion in a $15$s video in $4-8$ milliseconds on a GPU and a fraction of second on a CPU with model size of less than 1MB.

In another work, Hu et al. \cite{P1C5-2176} developed a 3D atrous CNN model to learn deep spatial-temporal features without losing resolution information. In addition, this model is combined with two convolutional long short-term memory (ConvLSTM) networks in order to capture both short-term and long-term spatio-temporal information of the input video data. Furthermore, 3D Atrous ConvLSTM
is a completely end-to-end framework that doesn't require any pre- or post-processing of the data. Experiments on CDnet 204 dataset show that 3D atrous CNN outperforms SuBSENSE \cite{P2C0-50}, Cascaded CNN \cite{P1C5-2110} and DeepBS \cite{P1C5-2140}. \\

\subsection{CNNs with Different Features}

\subsubsection{Random Permutation of Temporal Pixels (RPoTP) feature}
Zhao et al. \cite{P1C5-2182} designed a Deep Pixel Distribution Learning (DPDL) model for background subtraction. For the input of the CNNs, Zhao et al. \cite{P1C5-2182} employed a feature named  Random Permutation of Temporal Pixels (RPoTP) features instead of using the intensity values as in the previous methods. RPoTP is used to represent the distribution of past observations for a particular pixel, in which the temporal correlation between observations is deliberately no ordered over time. Then, a convolutional neural network (CNN) is used to learn the distribution for determining whether the current observation is foreground or background. The random permutation allows the framework to focus primarily on the distribution of observations, rather than be disturbed by spurious temporal correlations. For a large number of RPoTP features, the pixel representation is captured even with a small number of ground-truth frames. Experiments on the CDnet 2014 dataset show that DPDL is effective even with only a single ground-truth frame giving similar performance than the MOG model in this case. With $20$ GTs, DPDL obtains similar scores than SubSENSE \cite{P2C0-50}. Finally, DPDL\protect\footnotemark[26] with $40$ GTs gives an average F-Measure of $0.8106$ outperforming DeepBS \cite{P1C5-2140}. 

\footnotetext[26]{{hhttps://github.com/zhaochenqiu/DPDL}}

\subsubsection{Depth feature}
Wang et al. \cite{P1C5-2183} proposed a BackGround Subtraction neural Networks for Depth videos (BGSNet-D) to detect moving objects
in the scenarios where color information are unable to get. Thus, BGSNet-D is suitable in the dark scenes, where color information is hard to obtain.  CNNs can extract features in color images, but cannot applied to depth images directly because there exists edge noise and pixel absence in the captured data. To address this problem, Wang et al. \cite{P1C5-2183} designed an extended min-max normalization method to pre-process the depth images. After pre-processing, the two inputs of the CNNs are the average background image in depth and the current image in depth. Then the architecture is similar to ConvNets with three convolutional layers. In each
convolutional layer, a filter with $3 \times 3$ local receptive fields and  a $1 \times 1$ stride is used. ReLU follows as
the activation function in hidden layers. Batch normalization layer and pooling layer are after each ReLU layer. Finally, all feature maps are employed as inputs of a Multilayer Perceptron (MLP) which contains three fully connected layers. Sigmoid is used as activation
function and the output only consists of a single unit. Experiments on the SBM-RGBD\protect\footnotemark[27] dataset \cite{P6C2-Dataset-1026} show that BGSNet-D outperforms existing methods that use only depth data, and even reaches the performance of the methods that use RGB-D data.

\footnotetext[27]{{http://rgbd2017.na.icar.cnr.it/SBM-RGBDdataset.html}}

\subsection{Generative Adversarial Networks}
Bakkay et al. \cite{P1C5-2191} proposed a background subtraction method based on conditional Generative Adversarial Network (cGAN).
This model named BScGAN consists of two successive networks: generator and discriminator. The generator learns the mapping from the background and current image to the foreground mask. Then, the discriminator learns a loss function to train this mapping by comparing ground-truth and predicted output with observing the input image and background. For the architecture, the generator network follows an encoder-decoder architecture of Unet network with skip connections \cite{P1C5-2191-1}. Practically, the encoder part
includes down-sampling layers that decrease the size of the feature maps followed by convolutional filters. It consists of $8$ convolutional layers. The first layer uses $7 \times 7$ convolution to provide $64$ feature maps. The $8$th layer generates
$512$ feature maps with a $1 \times 1$ size. Their weights are randomly initialized. In addition, the middle $6$ convolutional
layers are six ResNet blocks. In all encoder layers, Leaky-ReLU non-linearities are used. For the decoder part, it uses upsampling layers followed by deconvolutional filters to construct an output image with the same resolution of the input one. Its architecture is similar to the encoder one including $8$ deconvolutional layers, but with a reverse layers ordering and with downsampling layers being replaced by up-sampling layers. For the discriminator network, the architecture is composed of $4$ convolutional and down-sampling layers. The first layer generates $64$ feature maps. Moreover, the $4$th layer generates $512$ feature maps with a $30 \times 30$ size. The convolutions are $3 \times 3$ spatial filters and their corresponding weights are randomly initialized. Leaky ReLU functions are employed as activation functions. Experimental results on CDnet 2014 datasets shows that BScGAN outperforms ConvNets \cite{P1C5-2100}, Cascaded CNN \cite{P1C5-2110}, and Deep CNN \cite{P1C5-2140} with an an average F-Measure of $0.9763$ without the category PTZ. 

Zheng et al. \cite{P1C5-2192} employed a Bayesian GAN (BGAN) approach. First, a median filter algorithm is used to extract the background and then a network based on Bayesian generative adversarial network is trained to classify each pixel, thereby dealing with the challenges of sudden and slow illumination changes, non-stationary background, and ghost. Practically, deep convolutional neural networks are adopted to construct the generator and the discriminator of Bayesian generative adversarial network. In a further work, Zheng et al. \cite{P1C5-2193} proposed a parallel version of the BGAN algorithm named (BPVGAN).

Bahri et al. \cite{P1C5-2194} designed an end-to-end framework called Neural Unsupervised Moving Object Detection (NUMOD). It 
is based on the batch method named ILISD \cite{P1C5-2194-1}. NUMOD can work either in an online and batch mode thanks to the parametrization via the generative neural network. NUMOD decomposes each frame into three parts: background, foreground and illumination changes. It uses a fully connected generative neural network to generate a background model by finding a low-dimensional manifold for the background of the image sequence. For the architecture, NUMOD uses two Generative Fully Connected Networks
(GFCN). Net1 estimates the background image from the input image while Net2 generates background image from
the illumination invariant image. These two networks have the exact same architecture. Thus, the input to GFCN is an optimizable low-dimensional latent vector. Then, two fully connected hidden layers are followed by ReLU non-linearity. The second hidden layer is fully connected to the output layer which is followed by the sigmoid function. A loss term is employed to impose the output of GFCN to be similar to the current input frame. Practically, GFCN is similar to the decoder part of an auto-encoder. In an auto-encoder, the low dimensional latent code is learned by the encoder, whilst in GFCN, it is a free parameter that can be optimized and is
the input to the network. During training, this latent vector learns a low-dimensional manifold of the input distribution. 

\section{Deep Learned Features}
\label{DeepLearnedFeatures}
Features used played an important role in the robustness against the challenge met in video \cite{P5C1-FS-32}. Historically, low-level and hand-craft features such as color \cite{P5C1-CF-203}\cite{P5C1-CF-230}, edge \cite{P5C1-EF-190}\cite{P5C1-EF-142}, texture \cite{P5C1-TF-11}\cite{P5C1-TF-63}, motion \cite{P5C1-MF-20}\cite{P5C1-MF-30}, and depth \cite{P5C1-SF-790}\cite{P5C1-SF-303}\cite{P5C1-SF-780}\cite{P5C1-SF-401}\cite{P5C1-SF-770}\cite{P5C1-SF-730} features were often employed to deal with illumination changes, dynamic background, and camouflage. But, it needs practically to choice an operator \cite{P5C1-FA-12}\cite{P5C1-FA-13}\cite{P5C1-FA-40} to fuse the results which come from the different features or a feature selection scheme \cite{P5C1-FS-30}\cite{P5C1-FS-31}. Nevertheless, none of these approaches can finally compete with approaches based on deep learned features. \\ 

\subsection{Stacked Denoising AutoEncoders}
Zhang et al. \cite{P1C5-2010} designed a deep learned features based block-wise method with a binary spatio-temporal background model. Based on the Stacked Denoising AutoEncoder (SDAE), the deep learning module learns a deep image representation encoding the intrinsic scene information. This leads to the robustness of feature description. Furthermore, the binary background model captures the spatio-temporal scene distribution information in the Hamming space to perform foreground detection. Experimental results \cite{P1C5-2010} on the CDnet 2012 dataset \cite{Dataset-2} demonstrate that SDAE gives better performance than traditional methods (MOG \cite{P1C2-MOG-10}, KDE \cite{P1C2-KDE-2}, LBP \cite{P5C1-TF-11}), and recent state-of-art model (PBAS \cite{P2C1-230}).   
To address robustness against stationary noise, Garcia-Gonzalez \cite{P1C5-2195} also used a stacked denoising autoencoders to generate a set of robust features for each patch of the image. Then, this set is considered as the input of a probabilistic model to determine if that region is background or foreground. \\

\subsubsection{Neural Reponse Mixture}
Shafiee et al. \cite{P1C5-2200}\cite{P1C5-2210} proposed a Neural Reponse Mixture (NeRM) framework to extract rich deep learned features with which to build a reliable MOG background model. Practically, the first synaptic layer of StochasticNet \cite{P1C5-2200-2} is trained on the ImageNet dataset \cite{P1C5-2200-1}  as a primitive, low-level, feature representation. Thus, the neural responses of the first synaptic layer at all pixels in the frame is then used as a feature to distinguish motion caused by objects moving in the scene. It is
worth noting that the formation of StochasticNets used in the NeRM framework is a one-time and off-line procedure which is not implemented on an embedded system. The final formed StochasticNet is transferred to the embedded system. Then, MOG model is employed  using the deep learned features. Experimental results  \cite{P1C5-2200} on the CDnet 2012 dataset \cite{Dataset-2} show that MOG-NeRM globally outperforms both the MOG model with RGB features and Color based Histogram model called CHist \cite{P1C5-2200-3}, but gives not the best score for the following categories: "intermittentObjectMotion"', "Low frame rate", "Night video", and "Thermal".  \\

\subsection{Motion Feature Network}
Nguyen et al. \cite{P1C5-2171} combined a sample-based background model with a feature extractor obtained by training a triplet network. This network  is constructed by three identical CNN, each of which is called a Motion Feature Network (MF-Net). Thus, each motion patterns is learned from small image patches and each input images of any size is transformed into feature embeddings for high-level representations. A sample based background model is then used with the color feature and the extracted deep motion features. To classify whether a pixel is background or foreground, Nguyen et al. \cite{P1C5-2171} employed the $l_1$ distance. Furthermore, an adaptive feedback scheme is also employed. The training is made with the CDNet 2014 dataset \cite{Dataset-1} and the offline trained network is then used on the fly without re-training on any video sequence before each execution. Experimental results \cite{P1C5-2171} on BMC 2012 dataset and CDNet 2014 dataset \cite{Dataset-1} show that MF-Net outperforms SOBS, LOBSTER and SuBSENSE in the case of dynamic backgrounds. Lee and Kim \cite{P1C5-2173} proposed a method to learn the pattern of the motions using the Factored 3-Way Restricted Boltzmann Machines (RBM) \cite{P1C5-2173-1} and obtain the global motion from the sequential images. Once this global motion is identified between frames, background subtraction is achieved by selecting the regions that do not respect the global motion. These regions are thus considered as the foreground region

\section{Adequacy for the background subtraction task}
\label{Discussion}
All the previous works demonstrated the performance of DNN for background subtraction but not discuss the reason why DNN works well. A first way to analyze these performance is to compare these different methods. For this, we have grouped in Table  \ref{BGS-Overview2} a comparative overview of the architectures while we show an overview in terms of the challenges in Table \ref{BGS-Overview3}. From Table \ref{BGS-Overview2}, we can see that it is possible to have three type of input: current image only, background and current images. In the first case, the authors works either with the current images without computing a background image or with a end-to-end solution that first generates a background image. In the second case, the authors have to compute the background image by using the temporal median or another model like SuBSENSE. The output is always the foreground mask except for NUMOD which provide the background and the foreground mask but also an illumination change mask. For the architecture, most of the authors employed a well-know architecture (LeNet-5, VGG-16 and U-Net) that they slighly adapted to the task of background subtraction. Only few authors proposed a full designed architecture for background subtraction. Table \ref{BGS-Overview3} groups the solutions of the different methods for the limitations of ConvNets \cite{P1C5-2100}. To learn the process at different level, the most common solutions are multi-scale and cascaded strategies alleviating the drawback to work with patches. For the training, over-fitting is often the case producing scene-specific methods. For the dataset used for the training, most of the authors employed the CDnet 2014 dataset with a part devoted to the training phase and another part for the testing phase. End-to-end solutions are well proposed as well as spatial and temporal strategies. Most of the time, the architecture is a generative one even if a combination of generative and discriminative would be better suitable for background subtraction. Indeed, the background modeling is more a reconstructive task while the foreground detection is more a discriminative task. 

To analyze how and why the DNN works well for this application, Minematsu et al. \cite{P1C5-2163}\cite{P1C5-2163-1} provided a valuable analysis by testing a quasi-similar method than ConvNet \cite{P1C5-2100} and found that the first layer performs the role of background subtraction using several filters whilst the last layer categorizes some background changes into a group without supervised signals. Thus, DNN automatically discovers background features through feature extraction by background subtraction and the integration of the features \cite{P1C5-2163} showing its potential for background/foreground separation. This first analysis is very valuable but the adequacy of a DNN method for the application of background/foreground separation should also be investigated in other key issues, that are the challenges and requirements met in background subtraction, and the adequacy of the architecture for background subtraction. 

To be effective, a background/foreground separation method should addressed the challenges and requirements met in this application, that are (1) its robustness to noise, (2) its spatial and temporal coherence, (3) the existence of an incremental version, (4) the existence of a real-time implementation, and (5) the ability to deal with the challenges met in video sequences. Practically, issue (1) is ensured for deep learning methods as DNN learn deep features of the background and the foreground during the training phase. For issue (2), spatial and temporal processing need to be added to pixel-wise DNN methods because, as explained in Alikan \cite{P1C5-2170}, one of the main challenges in DNN methods is dealing with objects of very different scales and the dithering effect at bordering pixels of foreground objects. In literature, several authors added spatial and temporal constraints via several spatial and/or temporal strategies. These  strategies can be either incorporated in an end-to-end solution or can be done via a post-processing applied to the foreground mask. For example,  Cascaded CNN \cite{P1C5-2110} and MV-FCN \cite{P1C5-2170} employed a multi-scale strategy while DeepBS \cite{P1C5-2140}  used a spatial median filter. Struct-CNN \cite{P1C5-2150} is based on a superpixel strategy whilst Attention ConvLSTM+CRF \cite{P1C5-2150} with Conditional Random Field (CRF). In another manner, Sakkos et al. \cite{P1C5-2167} used directly 3D-CNN for temporal coherence while Chen et al. \cite{P1C5-2166} used a spatial and temporal processing in Attention ConvLSTM. For issue (3), there is no need to update the background model in DNN methods if the training is sufficiently large to learn all the appearances of the model in terms of illumination changes and dynamics (waving trees, water rippling, waves, etc.), otherwise it is required. In this last case, several authors employed an end-to-end solution in which a DNN method for background generation is used to determine the background image over time. Then, the output of this DNN based  background generation is the input of the DNN based background subtraction with the current image  in order to determine the foreground mask.
For issue (4), DNNs are time consuming without a specific GPU card and optimizer. Thus, the key point to have a suitable DNN methods for background subtraction is to have a large training dataset, additional spatial/temporal strategies, and to apply it with a specific card if possible.  For issue (5) which concerns the challenges met in video sequences like illumination challenges and dynamic backgrounds, the DNN can be sufficient by itself if the architecture allow to learn these changes as in several works or  additional networks can be added.
 
For the adequacy of the architecture, it is needed to check the features of DNNs that are (1) type of architecture, and (2) parameters such as number of neurons, number of layers, etc. In literature, we can only found two works which compared different architecture for background/foreground separation: Cinelli \cite{P1C5-2100} tested both LeNet5 \cite{P1C5-2100-1} and ResNet \cite{P1C5-2100-2} architectures while  Chen et al. \cite{P1C5-2166} compared the VGG-16 \cite{P1C5-2150-2}, the GoogLeNet \cite{P1C5-2166-1}, and the ResNet50 \cite{P1C5-2100-2}. In these two works, ResNet \cite{P1C5-2100-2} provided the best results. But, these architectures were first designed for different classification tasks with the ImageNet dataset \cite{I-2}, CIFAR-10 dataset or ILSVRC 2015 dataset but not for the background/foreground separation task with the corresponding dataset such as CDnet 2014 dataset. \\

\section{Experimental results}
\label{sec:results}
For comparison, we present the results obtained on the well-known publicly available CDnet 2014 dataset \cite{Dataset-1} both in a qualitative and quantitative manner.

\subsection{CDnet 2014 dataset and Challenges}
\label{sec:Challenges}
CDnet 2014 dataset \cite{Dataset-1} was developed as part of Change Detection Workshop challenge (CDW 2014). This dataset includes all the videos from the CDnet 2012 dataset \cite{Dataset-2} plus 22 additional camera-captured videos providing 5 different categories that incorporate challenges that were not addressed in the 2012 dataset. Practically, the categories are as follows: baseline, dynamic backgrounds, camera jitter, shadows, intermittent object motion, thermal, challenging Weather, low frame-rate, night videos, PTZ and turbulence. In addition, whereas ground truths for all frames were made publicly available for the CDnet 2012 dataset for testing and evaluation, in the CDnet 2014, ground truths of only the first half of every video in the 5 new categories is made publicly available for testing. The evaluation will, however, be across all frames for all the videos (both new and old) as in CDnet 2012. All the challenges of these different categories have different spatial and temporal properties. It is important to determine what are the solved and unsolved challenges. Both CDnet 2012 and CDnet 2014 datasets allow to highlight in which situations it is difficult to provide robust foreground detection for existing background subtraction methods. The following remarks can be made as developed in \cite{P6C2-Dataset-1025}:
\begin{itemize}
\item Conventional background subtraction methods can efficiently deal with challenges met in baseline and bad weather sequences. 
\item Dynamic backgrounds, thermal video and camera jitter is a reachable challenge for top performing background subtraction. 
\item Night videos, low frame-rate, and PTZ videos represent huge challenges. 
\end{itemize}

\subsection{Performance Evaluation}
\subsubsection{Qualitative Evaluation}
\begin{figure} 
\centering 
\includegraphics[width=120mm]{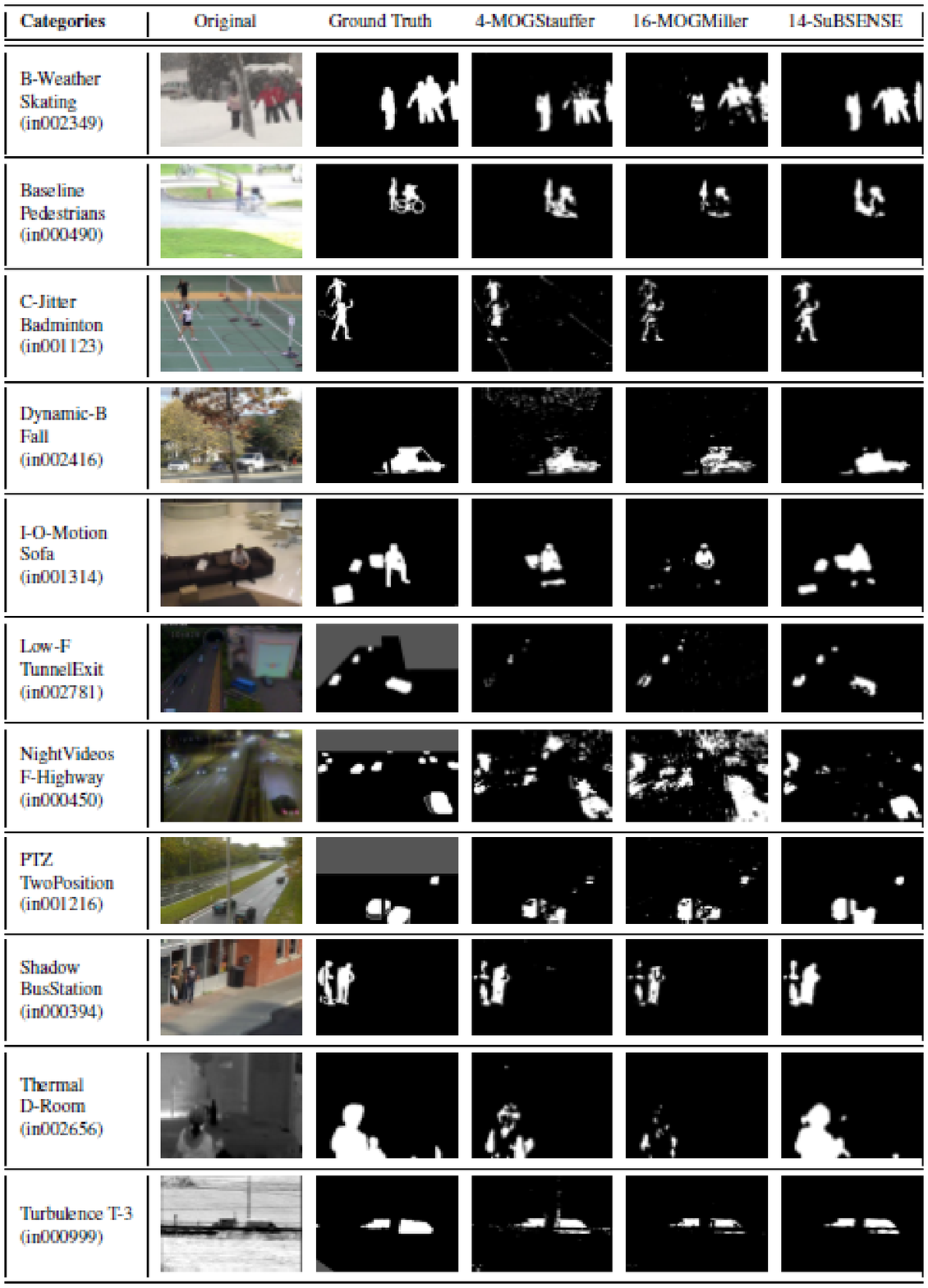} 
\caption{Visual results on CDnet 2014 dataset (Part 1): From left to right: Original images, Ground-Truth images, MOG (4-MOG-Stauffer \cite{P1C2-MOG-10}, RMOG (16-MOGMiller) \cite{P1C2-MOG-636}, SubSENSE \cite{P2C0-40}.} 
\label{Figure1}
\end{figure}

\begin{figure} 
\centering 
\includegraphics[width=120mm]{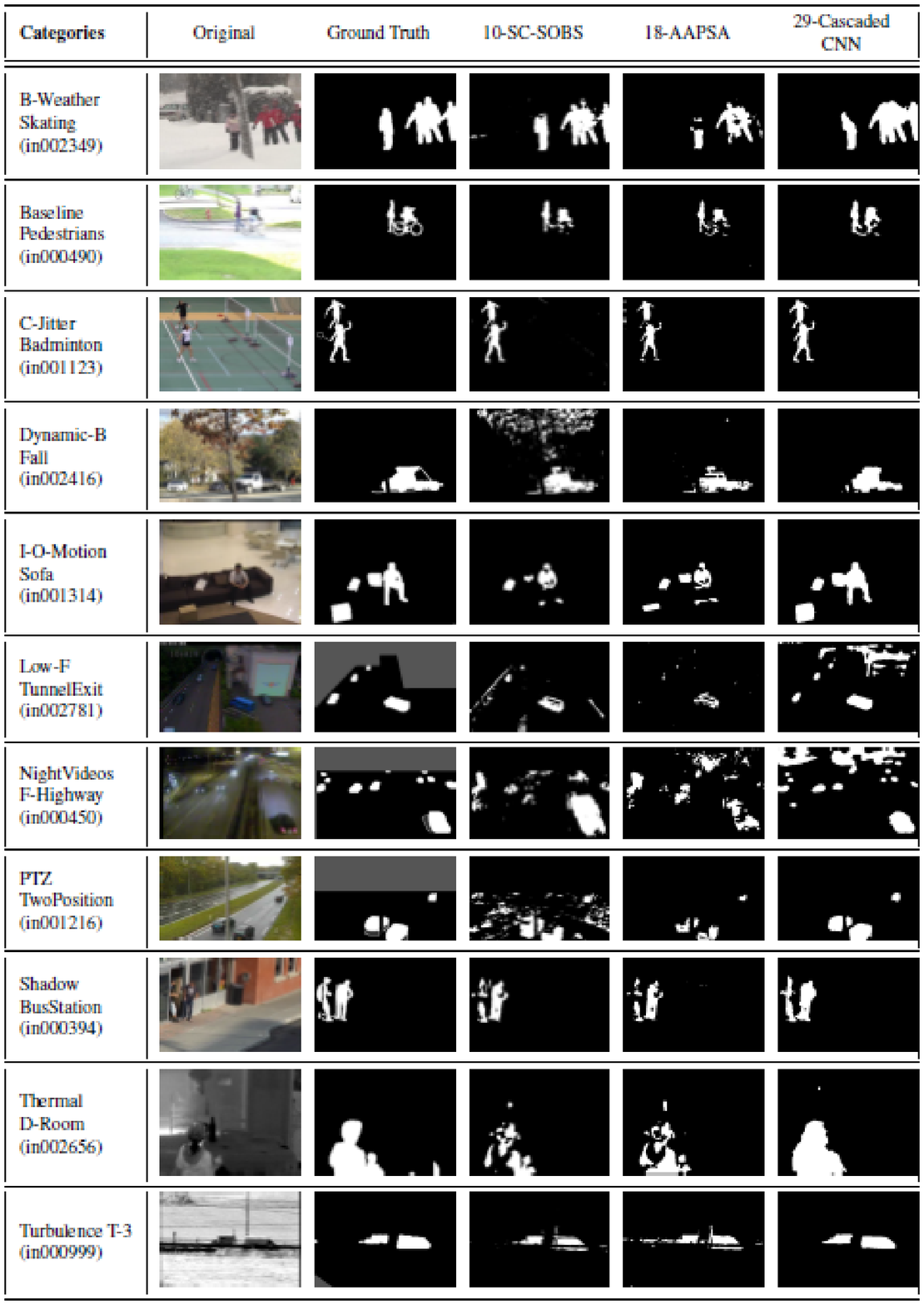} 
\caption{Visual results on CDnet 2014 dataset (Part 2): From left to right: Original images, Ground-Truth images, SC-SOBS \cite{P1C5-408}, AAPSA \cite{BI-6}, Cascaded CNN \cite{P1C5-2110}.} 
\label{Figure2}
\end{figure}

\begin{figure} 
\centering 
\includegraphics[width=120mm]{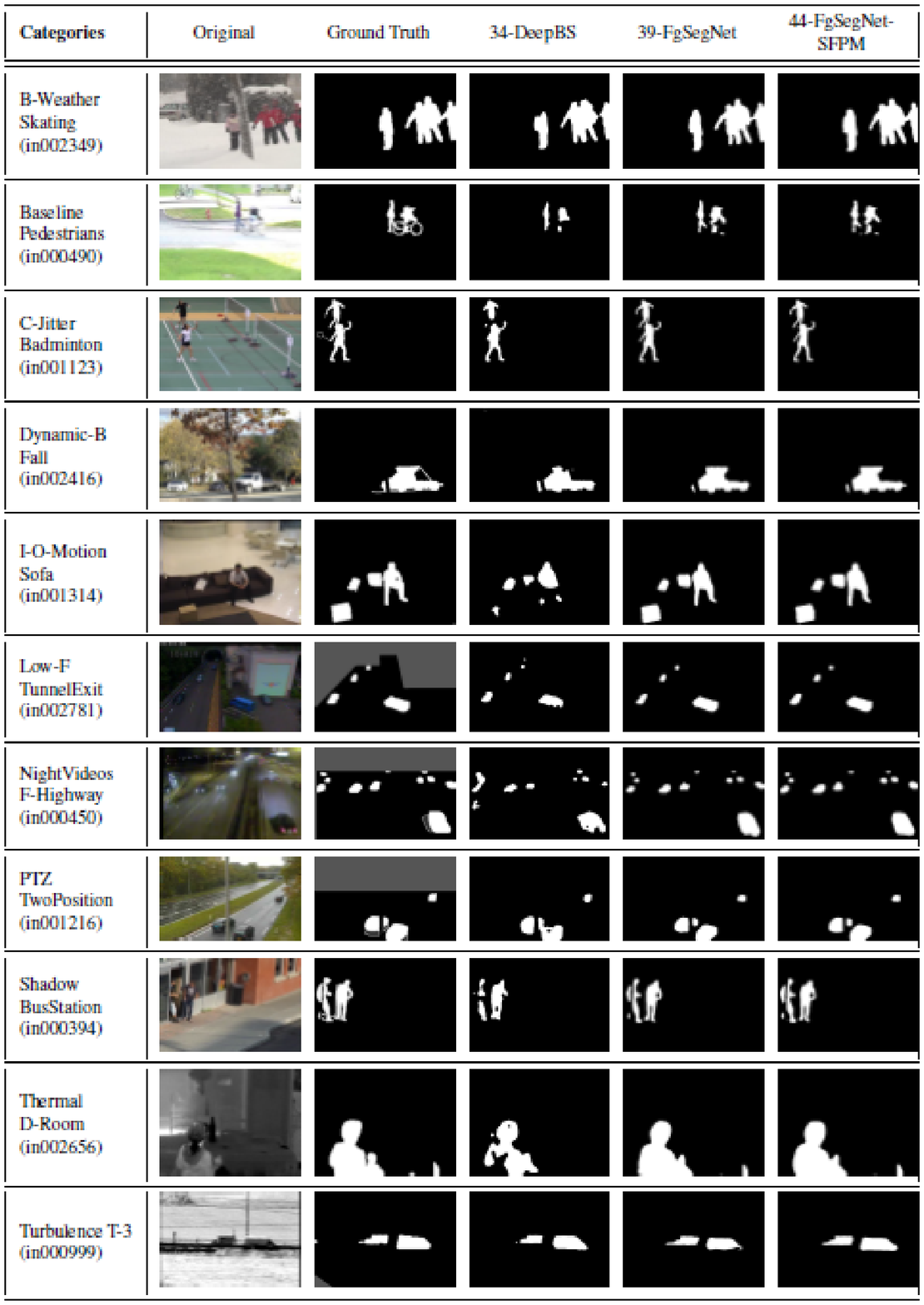} 
\caption{Visual results on CDnet 2014 dataset (Part 3): From left to right: Original images, Ground-Truth images, DeepBS \cite{P1C5-2140}, FgSegNet \cite{P1C5-2168}, FgSegNetSFPM \cite{P1C5-21680}.} 
\label{Figure3} 
\end{figure}

\begin{figure} 
\centering 
\includegraphics[width=120mm]{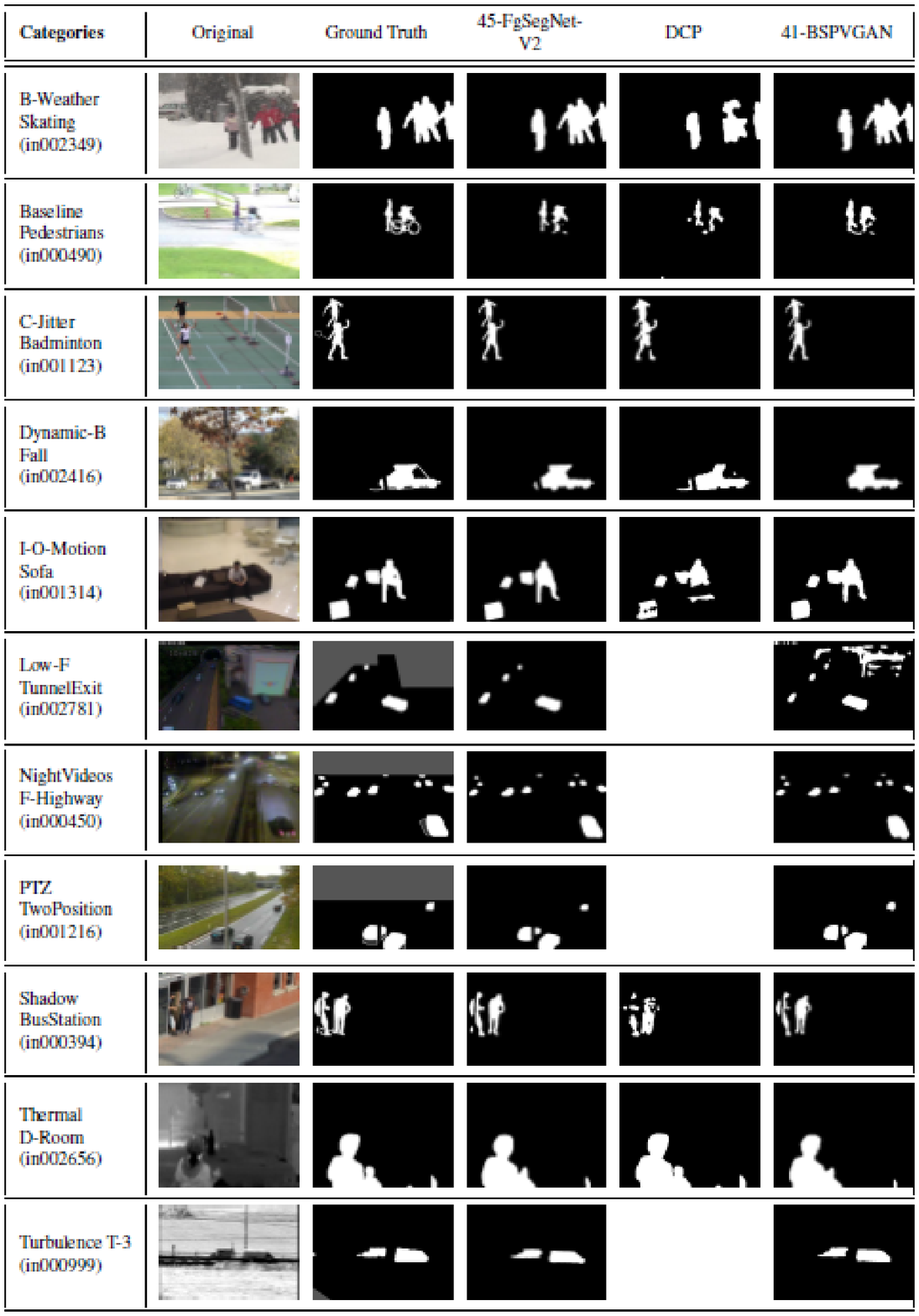} 
\caption{Visual results on CDnet 2014 dataset (Part 4): From left to right: Original images, Ground-Truth images, FgSegNet-V2 \cite{P1C5-21681}, DCP \cite{P1C5-2190}, BPVGAN \cite{P1C5-2193}. For DCP, the authors did not tested their algorithm on four categories.} 
\label{Figure4} 
\end{figure}

We compared the visual results obtained on the CDnet 2014 dataset by the different deep learning algorithms with visual results of other representative background subtraction algorithms that are: Two statistical models (MOG \cite{P1C2-MOG-10}, RMOG \cite{P1C2-MOG-636}), one multi-cues model (SubSENSE \cite{P2C0-40}), and two conventional neural networks (SC-SOBS \cite{P1C5-408}, AAPSA \cite{BI-6}). The deep learning models are the following ones: five CNNs based methods (Cascaded CNN \cite{P1C5-2110}, DeepBS \cite{P1C5-2140}, FgSegNet \cite{P1C5-2168}, FgSegNet-SFPM \cite{P1C5-21680}, FgSegNet-V2 \cite{P1C5-21681}) and two GANs based methods (BSPVGAN \cite{P1C5-2193}, DCP \cite{P1C5-2190}). All the visual results come from the CDnet 2014 website except for DCP for which the authors kindly provided the results. We also let in the four figures the number ID as well as the name as it is provided in the CDnet 2014 website. Figure \ref{Figure1} show the visual results obtained by MOG, RMOG  and SuBSENSE. We can see that SuBSENSE clearly improved the foreground mask by reducing false positive and negative detections. From Figure \ref{Figure2}, we can remark that Cascaded CNN outperforms the classical neural networks SC-SOBS and AAPSA except in the "Low-frame Rate" and "Night Videos" categories. In Figure \ref{Figure3}, FgSegNet and FgSegNet-SFPM that are top methods in CDnet 2014 dataset visually outperforms DeepBS in the "Baseline" and "Thermal"' Categories. In Figure \ref{Figure4}, FgSegNet-V2 which is the top method in CDnet 2014 dataset is compared with GAn based methods that give similar visual results. Finally, we can remark that the foreground mask was progressively improved over time by statistical models, multi-cue models, conventional neural networks, and deep learning models in the order of quality.

\subsubsection{Quantitative Evaluation}
We compared the F-measures obtained by the different algorithms with the F-measures of other representative background subtraction algorithms over the complete evaluation dataset: \textbf{(A)} two conventional statistical models (MOG \cite{P1C2-MOG-10}, RMOG \cite{P1C2-MOG-636}, \textbf{(B)} three advanced non-parametric models (SubSENSE \cite{P2C0-40}, PAWCS \cite{P2C0-50}, and Spectral-360 \cite{P5C1-CDFeatureI-500}), and \textbf{(C)} two conventional neural networks models (SOBS-CF \cite{P1C5-406}, SC-SOBS \cite{P1C5-408}). Deep learning models for background separation are classified following their architecture:
\begin{itemize}
\item \textbf{Convolutional Neural Networks:} We grouped scores of 20 algorithms based on CNN that are two basic CNN algorithms (two variants of ConvNet \cite{P1C5-2100}), six multi-scale or/and cascaded CNN algorithms (cascaded CNN \cite{P1C5-2110}, FgSegNet-M \cite{P1C5-2168}, FgSegNet-S \cite{P1C5-21680}, FgSegNet-V2 \cite{P1C5-21681}, MCSS \cite{P1C5-2179}, and Guided Multi-scale CNN \cite{P1C5-2177}), 1 fully CNN algorithms (MFCN \cite{P1C5-2174}), seven deep CNN algorithms (DeepBS  \cite{P1C5-2140},  TS-CNN \cite{P1C5-2161}, Joint TS-CNN \cite{P1C5-2161}, five variants of Attention ConvLSTM \cite{P1C5-2166}), one structured CNN algorithm (Struct-CNN \cite{P1C5-2150}), and two 3D CNN algorithms (3D CNN \cite{P1C5-2167}, 3D Atrous CNN \cite{P1C5-2176}).
\item \textbf{Generative Adversarial Networks:} We grouped scores of 4 GAN algorithms: DCP \cite{P1C5-2190}, BScGAN \cite{P1C5-2191}, BGAN \cite{P1C5-2192}, and BPVGAN \cite{P1C5-2193}.
\end{itemize}
Furthermore, these algorithms can be labeled as pixel-wise, spatial-wise, temporal-wise and spatio-temporal-wise algorithms. For pixel-wise algorithms, they were directly applied by the authors to background/foreground separation without specific processing taking into account spatial and temporal constraints. In these algorithms, each pixel is processed independently based or not on the information contained in their local patch like in ConvNet \cite{P1C5-2100}. Thus, they may produce isolated false positives and false negatives. For spatial-wise algorithms, these algorithms model the dependencies among adjacent spatial pixels and thus enforce spatial coherence like in Cascaded CNN \cite{P1C5-2110} and MFCN \cite{P1C5-2174} with a multi-scale strategy, Deep CNN (DeepBS) \cite{P1C5-2140} with spatial median filtering, Struct-CNN \cite{P1C5-2150} with super-pixel filtering, and Attention ConvLSTM+CRF \cite{P1C5-2150} with Conditional Random Field. For temporal-wise algorithms, these algorithms model the dependencies among adjacent temporal pixels and thus enforces temporal coherence such as Joint TS-CNN  \cite{P1C5-2161} with background reconstruction feedback and 3D-CNN \cite{P1C5-2167}. For spatio-temporal-wise algorithms, these algorithms model both the dependencies among adjacent spatial and temporal pixels and thus enforce both spatial and temporal coherence like Attention ConvLSTM+PSL+CRF \cite{P1C5-2166} with different architectures.Table \ref{ER1} groups the different F-measures which come either from the corresponding papers or directly from changedetection.net website. Barnich and Droogenbroeck \cite{P1C5-2100} did not test ConvNet on the Intermittent Motion Object (IOM) and PTZ categories because they claimed that their method is not designed for it. Similarly, Lim et al. \cite{P1C5-2150} did not evaluate Struct-CNN on the PTZ category as well as MCSS and BScGAN. Zeng and Zhu \cite{P1C5-2174} only evaluated MFCN on the THM category as this method is designed for infrared video. For those methods, the average F-Measure is done by indicating the missing category or the number of missing categories. For FgSegNet-M \cite{P1C5-2168}, FgSegNet-S \cite{P1C5-21680}, FgSegNet-V2 \cite{P1C5-21681}, we noticed that the F-Measure reported by the authors in their papers are different than the ones available on the CDnet website. We choose to report the one of the official CDnet, and the overal score provided by the authors are given between parenthesis. 
By analyzing Table \ref{ER1}, we can first see that the representative conventional neural networks Coherence-based and Fuzzy SOBS (SOBS-CF) \cite{P1C5-406} and SOBS with Spatial Coherence (SC-SOBS) \cite{P1C5-408} outperforms the basic statistical models like MOG \cite{P1C2-MOG-10} (1999) even with improvements like in RMOG \cite{P1C2-MOG-636} (2013). However, SOBS and its variants were the leader methods on the CDnet 2012 dataset \cite{Dataset-2} for a long time (around two years) showing the interest of neural networks for background subtraction. But, F-measure did not exceed $0.6$ in average, that were relatively low in absolute. The F-measure exceeded only $0,9$ for the baseline category making these methods only usable and reliable in applications where the environments were not too complex. Second, we can remark that advanced non parametric models as SuBSENSE \cite{P2C0-40} and PAWCS \cite{P2C0-50} developed in 2014/2015 achieved chronologically better performance than SOBS because of multi-features and multi-cues strategies. The gain in performance was around $25\%$ for the F-Measure. The average F-measure was around $0.75$ becoming to be more acceptable for a reliable use in real conditions especially that the F-measure was around $0.9$ for several challenges (baseline, dynamic backgrounds, camera jitter and shadow). Thus, these methods are more applicable in more complex environments. Third, we can observe that CNNs based method can achieve a maximum gap of performance around $30\%$ for the average F-Measure against SuBSENSE \cite{P2C0-40} and PAWCS \cite{P2C0-50} showing their superiority on this task. However, CNNs increase greatly the F-measure in the dynamic backgrounds, camera jitter, intermittent object motion and turbulence categories. For the PTZ category, the performance is mitigated as can be seen in works of several authors who did not provide results on this category arguing that they not designed their method for this challenge while score obtained by GANs are very interesting. Practically, these methods appear to be usable and reliable in a very large spectrum of environments, but there are most of the time scene-specific with a supervised mode. We can also see that the training has a great influence on the performance. Indeed, the results obtained by ConvNet using the manual foreground masks (GT) obtained a F-Measure around $0.9$ while this F-Measure falls around $0.79$ using the foreground masks from IUTIS showing in this case a little gap of performance in comparison with SuBSENSE \cite{P2C0-40} and PAWCS \cite{P2C0-50}. This fact also highlights that the gap of performance obtained by DNNs based methods is essentially due to their supervised aspects. In addition, their current computation times as can be seen in Table \ref{BGS-Overview3} are too slow to be currently employed in real applications.

\begin{landscape}
\begin{table*}[htb!]
\scalebox{0.70}{
\begin{tabular}{|l|l|l|l|l|l|l|l|l|l|l|l|l|}
\hline
\scriptsize{Algorithms (Authors)}  &\scriptsize{BSL}   &\scriptsize{DBG}   &\scriptsize{CJT}  
&\scriptsize{IOM}        &\scriptsize{SHD}   &\scriptsize{THM}   &\scriptsize{BDW}   &\scriptsize{LFR}   &\scriptsize{NVD}  
&\scriptsize{PTZ}        &\scriptsize{TBL}   &\scriptsize{Average} \\
\hline
\hline
\scriptsize{\textbf{A) Basic statistical models}}  &\scriptsize{}   &\scriptsize{}   &\scriptsize{}  
&\scriptsize{}        &\scriptsize{}   &\scriptsize{}   &\scriptsize{}   &\scriptsize{}   &\scriptsize{}  
&\scriptsize{}        &\scriptsize{}  &\scriptsize{} \\

\scriptsize{MOG  (Stauffer and Grimson \cite{P1C2-MOG-10} 1999)}   &\scriptsize{0.8245}  &\scriptsize{0.6330} &\scriptsize{0.5969} 
																				 &\scriptsize{0.5207}  &\scriptsize{0.7156} &\scriptsize{0.6621} 
																				 &\scriptsize{0.7380}  &\scriptsize{0.5373} &\scriptsize{0.4097} 
																				 &\scriptsize{0.1522}  &\scriptsize{0.4663} &\scriptsize{0.5707} \\															
\scriptsize{RMOG  (Varadarajan et al. \cite{P1C2-MOG-636} 2013)} &\scriptsize{0.7848}  &\scriptsize{0.7352} &\scriptsize{0.7010} 
																				 &\scriptsize{0.5431}  &\scriptsize{0.7212} &\scriptsize{0.4788}
																				 &\scriptsize{0.6826}  &\scriptsize{0.5312} &\scriptsize{0.4265} 
																				 &\scriptsize{0.2400}  &\scriptsize{0.4578} &\scriptsize{0.5735} \\
\hline
\scriptsize{\textbf{B) Advanced non parametric models}}     &\scriptsize{}   &\scriptsize{}   &\scriptsize{}  
&\scriptsize{}        &\scriptsize{}   &\scriptsize{}   &\scriptsize{}   &\scriptsize{}   &\scriptsize{}  
&\scriptsize{}        &\scriptsize{}   &\scriptsize{} \\																		
\hline
\scriptsize{SuBSENSE (St-Charles et al. \cite{P2C0-40} 2014)}   &\scriptsize{0.9503}  &\scriptsize{0.8117} &\scriptsize{0.8152} 
																				 &\scriptsize{0.6569}  &\scriptsize{0.8986} &\scriptsize{0.8171} 
																				 &\scriptsize{0.8619}  &\scriptsize{0.6445} &\scriptsize{0.5599} 
																				 &\scriptsize{0.3476}  &\scriptsize{0.7792} &\scriptsize{0.7408} \\																				
\scriptsize{PAWCS  (St-Charles et al. \cite{P2C0-50} 2015)}     &\scriptsize{0.9397}  &\scriptsize{0.8938} &\scriptsize{0.8137} 
																				 &\scriptsize{0.7764}  &\scriptsize{0.8913} &\scriptsize{0.8324} 
																				 &\scriptsize{0.8152}  &\scriptsize{0.6588} &\scriptsize{0.4152} 
																				 &\scriptsize{0.4615}  &\scriptsize{0.6450} &\scriptsize{0.7403} \\ 
\scriptsize{Spectral-360 (Sedky et al. \cite{P5C1-CDFeatureI-500} 2014)} &\scriptsize{0.9330}  &\scriptsize{0.7872} &\scriptsize{0.7156} 
																											 &\scriptsize{0.5656}  &\scriptsize{0.8843} &\scriptsize{0.7764}
																											 &\scriptsize{0.7569}  &\scriptsize{0.6437} &\scriptsize{0.4832} 
																											 &\scriptsize{0.3653}  &\scriptsize{0.5429} &\scriptsize{0.7054} \\
\hline																											
\scriptsize{\textbf{C) Conventional Neural Networks}}  &\scriptsize{}   &\scriptsize{}   &\scriptsize{}  
&\scriptsize{}        &\scriptsize{}   &\scriptsize{}   &\scriptsize{}   &\scriptsize{}   &\scriptsize{}  
&\scriptsize{}        &\scriptsize{}  &\scriptsize{} \\
\scriptsize{SOBS-CF  (Maddalena and Petrosino \cite{P1C5-406} 2010)}   &\scriptsize{0.9299}  &\scriptsize{0.6519} &\scriptsize{0.7150} 
																				 &\scriptsize{0.5810}  &\scriptsize{0.7045} &\scriptsize{0.7140} 
																				 &\scriptsize{0.6370}  &\scriptsize{0.5148} &\scriptsize{0.4482} 
																				 &\scriptsize{0.0368}  &\scriptsize{0.4702} &\scriptsize{0.5883} \\
\scriptsize{SC-SOBS  (Maddalena and Petrosino \cite{P1C5-408} 2012)}   &\scriptsize{0.9333}  &\scriptsize{0.6686} &\scriptsize{0.7051} 
																				 &\scriptsize{0.5918}  &\scriptsize{0.7230} &\scriptsize{0.6923} 
																				 &\scriptsize{0.6620}  &\scriptsize{0.5463} &\scriptsize{0.4503} 
																				 &\scriptsize{0.0409}  &\scriptsize{0.4880} &\scriptsize{0.5961} \\
\hline																										
\scriptsize{\textbf{D) Deep Neural Networks} (\textit{Structure})}               &\scriptsize{}   &\scriptsize{}   &\scriptsize{}  
&\scriptsize{}        &\scriptsize{}   &\scriptsize{}   &\scriptsize{}   &\scriptsize{}   &\scriptsize{}  
&\scriptsize{}        &\scriptsize{}   &\scriptsize{} \\	
\hline		
\scriptsize{\textbf{1) Convolutional Neural Networks}} &\scriptsize{}   &\scriptsize{}   &\scriptsize{}  
&\scriptsize{}        &\scriptsize{}   &\scriptsize{}    &\scriptsize{}   &\scriptsize{}   &\scriptsize{}  
&\scriptsize{}        &\scriptsize{}   &\scriptsize{} \\	
\hline			
\scriptsize{\textbf{1.1) Basic CNN}} &\scriptsize{}   &\scriptsize{}   &\scriptsize{}  
&\scriptsize{}        &\scriptsize{}   &\scriptsize{}    &\scriptsize{}   &\scriptsize{}   &\scriptsize{}  
&\scriptsize{}        &\scriptsize{}   &\scriptsize{} \\	
\scriptsize{CNN$^*$ (ConvNet-GT) (\textit{LeNet-5}) (\textit{Pixel-wise}) (Barnich and Droogenbroeck \cite{P1C5-2100} 2016)}       
																								&\scriptsize{0.9813}  &\scriptsize{0.8845} &\scriptsize{0.9020} 	
																								&\scriptsize{-}       &\scriptsize{0.9454} &\scriptsize{0.8543} 
                                                &\scriptsize{0.9254}  &\scriptsize{0.9612} &\scriptsize{0.7565}		
																								&\scriptsize{-}     &\scriptsize{0.9297}   &\scriptsize{0.9044 (IOM, PTZ)} \\
\scriptsize{CNN$^*$ (ConvNet-IUTIS) (\textit{LeNet-5}) (\textit{Pixel-wise}) (Barnich and Droogenbroeck \cite{P1C5-2100} 2016)} &\scriptsize{0.9647} &\scriptsize{0.7923} &\scriptsize{0.8013} 
																											 &\scriptsize{-}&\scriptsize{0.8590} &\scriptsize{0.7559} 
																											 &\scriptsize{0.8849}  &\scriptsize{0.8273} &\scriptsize{0.4715} 
																											 &\scriptsize{-}&\scriptsize{0.7506}   &\scriptsize{0.7897 (IOM, PTZ)} \\
\scriptsize{$DPDL_{1}$$^*$ (One GT) (\textit{CNN}) (\textit{Temporal-wise}) (Zhao et al. \cite{P1C5-2182} 2018)} &\scriptsize{0.7886} &\scriptsize{0.6566} &\scriptsize{0.5456} 
																											 &\scriptsize{0.5115} &\scriptsize{0.6957} &\scriptsize{0.6697} 
																											 &\scriptsize{0.6036}  &\scriptsize{0.5966} &\scriptsize{0.3953} 
																											 &\scriptsize{0.2942}&\scriptsize{0.6301}   &\scriptsize{0.5807} \\
\scriptsize{$DPDL_{20}$$^*$ (20 GTs) (\textit{CNN}) (\textit{Temporal-wise}) (Zhao et al. \cite{P1C5-2182} 2018)} &\scriptsize{0.9620} &\scriptsize{0.8369} &\scriptsize{0.8627} 
																											 &\scriptsize{0.8174} &\scriptsize{0.8763} &\scriptsize{0.8311} 
																											 &\scriptsize{0.8107}  &\scriptsize{0.6646} &\scriptsize{0.5866}
																											 &\scriptsize{0.4654} &\scriptsize{0.7173}   &\scriptsize{0.7665} \\																										\scriptsize{$DPDL_{40}$$^*$ (40GT) (\textit{CNN}) (\textit{Temporal-wise}) (Zhao et al. \cite{P1C5-2182} 2018)} &\scriptsize{0.9692} &\scriptsize{0.8692} &\scriptsize{0.8661} 
																											 &\scriptsize{0.8759} &\scriptsize{0.9361} &\scriptsize{0.8379} 
																											 &\scriptsize{0.8688}  &\scriptsize{0.7078} &\scriptsize{0.6110} 
																											 &\scriptsize{0.6087} &\scriptsize{0.7636}   &\scriptsize{0.8106} \\					\hline
\scriptsize{\textbf{1.2) Multi-scale or/and Cascaded CNNs}} &\scriptsize{}   &\scriptsize{}   &\scriptsize{}  
&\scriptsize{}        &\scriptsize{}   &\scriptsize{}    &\scriptsize{}   &\scriptsize{}   &\scriptsize{}  
&\scriptsize{}        &\scriptsize{}   &\scriptsize{} \\																							
\scriptsize{Cascaded CNN (\textit{CNN-1/CNN-2}) (\textit{Spatial-wise}) (Wang et al. \cite{P1C5-2110} 2016)} &\scriptsize{0.9786}  &\scriptsize{0.9658} &\scriptsize{0.9758} 
																											  &\scriptsize{0.8505}  &\scriptsize{0.9414} &\scriptsize{0.8958}
																											  &\scriptsize{0.9431}  &\scriptsize{0.8370} &\scriptsize{0.8965} 
	   																			              &\scriptsize{0.9168}  &\scriptsize{0.9108} &\scriptsize{0.9209}  \\
\scriptsize{FgSegNet-M (\textit{-}) (\textit{Spatial-wise})  (Lim and Keles \cite{P1C5-2168} 2018)} &\scriptsize{0.9973}  &\scriptsize{0.9958} &\scriptsize{0.9954} 
																											               &\scriptsize{0.9951}  &\scriptsize{0.9937} &\scriptsize{0.9921}
																											               &\scriptsize{0.9845}  &\scriptsize{0.8786} &\scriptsize{0.9655} 
																																	   &\scriptsize{0.9843}  &\scriptsize{0.9648} &\scriptsize{0.9770 (0.9865$^*$)} \\
																																		
\scriptsize{FgSegNet-S (\textit{-}) (\textit{Spatial-wise})  (Lim and Keles \cite{P1C5-21680} 2018)} &\scriptsize{0.9977}  &\scriptsize{0.9958} 
&\scriptsize{0.9957} 
																											               &\scriptsize{0.9940}  &\scriptsize{0.9927} &\scriptsize{0.9937}
																											               &\scriptsize{0.9897}  &\scriptsize{0.8972} &\scriptsize{0.9713} 
																																	   &\scriptsize{0.9879}  &\scriptsize{0.9681} &\scriptsize{0.9804 (0.9878$^*$)} \\
																																		
\scriptsize{FgSegNet-V2 (\textit{-}) (\textit{Spatial-wise})  (Lim et al. \cite{P1C5-21681} 2018)} &\scriptsize{0.9978}  &\scriptsize{0.9951} &\scriptsize{0.9938} 
																											               &\scriptsize{0.9961}  &\scriptsize{0.9955} &\scriptsize{0.9938}
																											               &\scriptsize{0.9904}  &\scriptsize{0.9336} &\scriptsize{0.9739} 
																																	   &\scriptsize{0.9862}  &\scriptsize{0.9727} &\scriptsize{0.9847 (0.9890$^*$)} \\
\scriptsize{MCSS$^*$ (\textit{-})  (\textit{Spatial-wise}) (Liao et al. \cite{P1C5-2179} 2018)} &\scriptsize{0.9940}  &\scriptsize{0.881} &\scriptsize{0.794} 
																											               &\scriptsize{0.770}  &\scriptsize{0.915} &\scriptsize{0.883}
																											               &\scriptsize{0.861}  &\scriptsize{0.725} &\scriptsize{0.788} 
																																	   &\scriptsize{-}  &\scriptsize{0.884} &\scriptsize{0.844} \\
\scriptsize{Guided Multi-scale CNN$^*$ (\textit{-})  (\textit{Spatial-wise}) (Liang et al. \cite{P1C5-2177} 2018)} &\scriptsize{0.9791}  
&\scriptsize{0.8266} &\scriptsize{0.8818} 
&\scriptsize{0.6229}  &\scriptsize{0.8910} &\scriptsize{0.7490}
																											             &\scriptsize{0.8711}  &\scriptsize{0.6396} &\scriptsize{0.5048} 
																																	 &\scriptsize{0.6057}  &\scriptsize{0.8114} &\scriptsize{0.7591} \\																																	
\hline																																																				
\scriptsize{\textbf{1.3) Fully CNNs}} &\scriptsize{}   &\scriptsize{}   &\scriptsize{}  
&\scriptsize{}        &\scriptsize{}   &\scriptsize{}    &\scriptsize{}   &\scriptsize{}    &\scriptsize{}  &\scriptsize{}  &\scriptsize{}   &\scriptsize{} \\ 
\scriptsize{MFCN (\textit{-}) (\textit{Spatial-wise}) (Zeng and Zhu \cite{P1C5-2174} 2018)}   &\scriptsize{-}  &\scriptsize{-} 
&\scriptsize{-} 
																											  &\scriptsize{-}  &\scriptsize{-} &\scriptsize{0.9870}
																											  &\scriptsize{-}  &\scriptsize{-} &\scriptsize{-} 
	   																			              &\scriptsize{-}  &\scriptsize{-} &\scriptsize{0.9870 (only THM)}  \\	
\hline
\scriptsize{\textbf{1.4) Deep CNNs}} &\scriptsize{}   &\scriptsize{}   &\scriptsize{}  
&\scriptsize{}        &\scriptsize{}   &\scriptsize{}    &\scriptsize{}   &\scriptsize{}   &\scriptsize{}  
&\scriptsize{}        &\scriptsize{}   &\scriptsize{} \\
\scriptsize{Deep CNN (DeepBS) (\textit{-}) (\textit{Pixel-wise}) (Babaee et al. \cite{P1C5-2140} 2017)}   &\scriptsize{0.9580}  &\scriptsize{0.8761} &\scriptsize{0.8990} 
																											 &\scriptsize{0.6098}  &\scriptsize{0.9304} &\scriptsize{0.7583}
																											 &\scriptsize{0.8301}  &\scriptsize{0.6002} &\scriptsize{0.5835} 
	   																			             &\scriptsize{0.3133}  &\scriptsize{0.8455} &\scriptsize{0.7548} \\	
\scriptsize{Two-Stage CNN$^*$ (TS-CNN) (\textit{-}) (\textit{Pixel-wise})  (Zhao et al. \cite{P1C5-2161} 2018)} &\scriptsize{0.9630}  &\scriptsize{0.7405} 
&\scriptsize{0.8689} 
																											  &\scriptsize{0.8734}  &\scriptsize{0.9216} &\scriptsize{0.8536}
																											  &\scriptsize{0.8004}  &\scriptsize{0.8075} &\scriptsize{0.6851} 
	   																			              &\scriptsize{0.4493}  &\scriptsize{0.6929} &\scriptsize{0.7870}   \\	
\scriptsize{Joint TS-CNN$^*$ (\textit{-}) (\textit{Temporal-wise}) (Zhao et al. \cite{P1C5-2161} 2017)} &\scriptsize{0.9680}  &\scriptsize{0.7716} &\scriptsize{0.8988} 
																											  &\scriptsize{0.9066}  &\scriptsize{0.9286} &\scriptsize{0.8586}
																											  &\scriptsize{0.8550}  &\scriptsize{0.7491} &\scriptsize{0.7695} 
	   																			              &\scriptsize{0.5168}  &\scriptsize{0.7143} &\scriptsize{0.8124}   \\
\scriptsize{Attention ConvLSTM$^*$ (\textit{VGG-16}) (\textit{Temporal-wise}) (Chen et al. \cite{P1C5-2166} 2018)} &\scriptsize{0.9243}  &\scriptsize{0.6030} 
&\scriptsize{0.9053} 
																											          &\scriptsize{0.572}  &\scriptsize{0.8916} &\scriptsize{0.7181}
																											          &\scriptsize{0.8493}  &\scriptsize{0.5920} &\scriptsize{0.5060} 
	   																			                      &\scriptsize{0.7436}  &\scriptsize{0.7347} &\scriptsize{0.7314} \\
\scriptsize{Attention ConvLSTM+CRF$^*$ (\textit{VGG-16})  (\textit{Spatial/Temporal-wise})(Chen et al. \cite{P1C5-2166} 2018)} &\scriptsize{0.9383}  &\scriptsize{0.6207} 
&\scriptsize{0.9251} 
																											              &\scriptsize{0.6058}  &\scriptsize{0.8962} &\scriptsize{0.7271}
																											              &\scriptsize{0.8846}  &\scriptsize{0.6113} &\scriptsize{0.5188} 
	   																			                          &\scriptsize{0.7697}  &\scriptsize{0.7404} &\scriptsize{0.7489} \\
\scriptsize{Attention ConvLSTM+PSL+CRF$^*$ (\textit{VGG-16}) (\textit{Spatial/Temporal-wise}) (Chen et al. \cite{P1C5-2166} 2018)} &\scriptsize{0.9594}  &\scriptsize{0.7356} 
&\scriptsize{0.9422} 
																											            &\scriptsize{0.7538}  &\scriptsize{0.9084} &\scriptsize{0.8546}
																											            &\scriptsize{0.8949}  &\scriptsize{0.6175} &\scriptsize{0.7526} 
	   																			                        &\scriptsize{0.7816}  &\scriptsize{0.9207} &\scriptsize{0.8292} \\
\scriptsize{Attention ConvLSTM+PSL+CRF$^*$ (\textit{GoogleLeNet}) (\textit{Spatial/Temporal-wise}) (Chen et al. \cite{P1C5-2166} 2018)} &\scriptsize{0.8557} 
&\scriptsize{0.6588} &\scriptsize{0.8864} 
																											            &\scriptsize{0.6488}  &\scriptsize{0.8049} &\scriptsize{0.7725}
																											            &\scriptsize{0.7961}  &\scriptsize{0.5947} &\scriptsize{0.6003} 
	   																			                        &\scriptsize{0.7136}  &\scriptsize{0.7637} &\scriptsize{0.7360} \\
\scriptsize{Attention ConvLSTM+PSL+CRF$^*$ (\textit{ResNet}) (\textit{Spatial/Temporal-wise}) (Chen et al. \cite{P1C5-2166} 2018)} &\scriptsize{0.9294}  &\scriptsize{0.8220} &\scriptsize{0.9518} 
																											            &\scriptsize{0.8453}  &\scriptsize{0.9647} &\scriptsize{0.9444}
																											            &\scriptsize{0.9461}  &\scriptsize{0.8080} &\scriptsize{0.8585} 
	   																			                        &\scriptsize{0.7776}  &\scriptsize{0.8011} &\scriptsize{0.8772} \\
																												
\hline
\scriptsize{\textbf{1.5) Structured CNNs}} &\scriptsize{}   &\scriptsize{}   &\scriptsize{}  
&\scriptsize{}        &\scriptsize{}   &\scriptsize{}    &\scriptsize{}   &\scriptsize{}   &\scriptsize{} &\scriptsize{}  &\scriptsize{} &\scriptsize{} \\
\scriptsize{Struct-CNN$^*$ (\textit{VGG-16}) (\textit{Spatial-wise}) (Lim et al.	\cite{P1C5-2150} 2017)} &\scriptsize{0.9586}  &\scriptsize{0.9112} &\scriptsize{0.8990} 
																											  &\scriptsize{0.8780}  &\scriptsize{0.8565} &\scriptsize{0.8048}
																											  &\scriptsize{0.8757}  &\scriptsize{0.9321} &\scriptsize{0.7715} 
	   																			              &\scriptsize{-}  &\scriptsize{0.7573} &\scriptsize{0.8645}  \\
\hline
\scriptsize{\textbf{1.6) 3D CNNs}} &\scriptsize{}   &\scriptsize{}   &\scriptsize{}  
&\scriptsize{}        &\scriptsize{}   &\scriptsize{}    &\scriptsize{}   &\scriptsize{}   &\scriptsize{}   &\scriptsize{} &\scriptsize{} &\scriptsize{}  \\					
\scriptsize{3D CNN$^*$ (\textit{C3D branch}) (\textit{Temporal-wise}) (Sakkos et al. \cite{P1C5-2167} 2017)} &\scriptsize{0.9691}  &\scriptsize{0.9614}
 &\scriptsize{0.9396} 
																											            &\scriptsize{0.9698}  &\scriptsize{0.9706} &\scriptsize{0.9830}
																											            &\scriptsize{0.9509}  &\scriptsize{0.8862} &\scriptsize{0.8565} 
	   																			                        &\scriptsize{0.8987}  &\scriptsize{0.8823} &\scriptsize{0.9507} \\
\scriptsize{3D Atrous CNN$^*$ (ConvLTSM) (\textit{-}) (\textit{Spatial/Temporal-wise}) (Hu et al. \cite{P1C5-2176} 2018)} &\scriptsize{0.9897}  &\scriptsize{0.9789}
 &\scriptsize{0.9645} 
																											            &\scriptsize{0.9637}  &\scriptsize{0.9813} &\scriptsize{0.9833}
																											            &\scriptsize{0.9609}  &\scriptsize{0.8994} &\scriptsize{0.9489} 
	   																			                        &\scriptsize{0.8582}  &\scriptsize{0.9488} &\scriptsize{0.9615} \\	
\hline
\scriptsize{\textbf{2) Generative Adversarial Networks}}    &\scriptsize{}   &\scriptsize{}   &\scriptsize{}  
&\scriptsize{}        &\scriptsize{}   &\scriptsize{}   &\scriptsize{}   &\scriptsize{}   &\scriptsize{}  
&\scriptsize{}        &\scriptsize{}   &\scriptsize{} \\	
\scriptsize{DCP$^*$ (\textit{VGG-19}) (Sultana et al. \cite{P1C5-2190} 2018)}   &\scriptsize{0.8178}   &\scriptsize{0.7757}   &\scriptsize{0.8376}  
																													&\scriptsize{0.5979}    &\scriptsize{0.7665}   &\scriptsize{0.8212}   
																													&\scriptsize{0.8212}   &\scriptsize{-}   &\scriptsize{-}  
																													&\scriptsize{-}        &\scriptsize{-}   &\scriptsize{0.7620 (4)}\\	
\scriptsize{BScGAN$^*$  (\textit{UNet/ResNet})  (\textit{Pixel-wise}) (Bakkay et al. \cite{P1C5-2191} 2018)}  &\scriptsize{0.9930}  &\scriptsize{0.9784} &\scriptsize{0.9770} 
																					 &\scriptsize{0.9623}  &\scriptsize{0.9828} &\scriptsize{0.9612} 
																					 &\scriptsize{0.9796}  &\scriptsize{0.9918} &\scriptsize{0.9661} 
																					 &\scriptsize{-}  &\scriptsize{0.9712} &\scriptsize{0.9763 (PTZ)} \\
																					
\scriptsize{BGAN  (\textit{-}) (\textit{Pixel-wise}) (Zheng et al. \cite{P1C5-2192} 2018)}  &\scriptsize{0.9814}  &\scriptsize{0.9763} &\scriptsize{0.9828} 
																					 &\scriptsize{0.9366}  &\scriptsize{0.9849} &\scriptsize{0.9064} 
																					 &\scriptsize{0.9465}  &\scriptsize{0.8472} &\scriptsize{0.8965} 
																					 &\scriptsize{0.9194}  &\scriptsize{0.9118} &\scriptsize{0.9339} \\
																				
\scriptsize{BPVGAN  (\textit{-}) (\textit{Pixel-wise}) (Zheng et al. \cite{P1C5-2193} 2018)}  &\scriptsize{0.9837}  &\scriptsize{0.9849} &\scriptsize{0.9893} 
																					 &\scriptsize{0.9366}  &\scriptsize{0.9927} &\scriptsize{0.9764} 
																					 &\scriptsize{0.9644}  &\scriptsize{0.8508} &\scriptsize{0.9001} 
																					 &\scriptsize{0.9486}  &\scriptsize{0.9310} &\scriptsize{0.9501} \\
\hline
\end{tabular}}
\caption{F-measure metric over the 6 categories of the CDnet2014, namely Baseline (BSL), Dynamic background (DBG), Camera jitter (CJT)Intermittent Motion Object (IOM), Shadows (SHD), Thermal (THM), Bad Weather (BDW), Low Frame Rate (LFR), Night Videos (NVD), PTZ, Turbulence (TBL). $^*$ indicated that the measures come from the corresponding papers otherwise the measures comes from the changedetection.net website.}
\centering
\label{ER1}
\end{table*}
\end{landscape}

\section{Conclusion}
\label{sec:Conclusion}
In this paper, we have firstly presented a full review of recent advances on deep neural networks applied to background generation, background subtraction and deep learned features for detection of moving objects in video taken by a static camera. Experimental results on the large-scale CDnet 2014 dataset show the gap of performance obtained by the supervised deep neural networks methods in this field. Even if deep neural networks has received significant attention much more recently for background subtraction in the last two years since the seminal paper of Braham and Van Droogenbroeck \cite{P1C5-2100}, there are many unsolved important issues:
\begin{itemize}
\item The main question is what is the best suitable type of deep neural networks and its corresponding architecture for background initialization, background subtraction and deep learned features in presence of complex backgrounds? 
\item Looking at the experiments, several authors avoid experiments on the "PTZ" category and when the F-Measure is provided the score is not always very high. Thus, it seems that the current deep neural networks tested meet problems in the case of moving cameras. 
\item For the inputs, all the authors employed either gray or color images in RGB, except \cite{P1C5-2182} which used a distribution learning feature improving the performance of the basic CNNs. But, it would be surely interesting to employ RGB-D images because depth information is very helpful in several challenges like camouflage as developed in Maddalena and Petrosino \cite{Survey-11}. In addition, the conventional neural networks SOBS \cite{P1C5-410-1} is the top algorithm on the SBM-RGBD dataset \cite{P6C2-Dataset-1026}. Thus, we can expect that CNNs with RGB-D features as inputs will also achieve great performance as ForeGAN-RGBD \cite{P1C5-2190-1}model. However, multi-spectral data would be also interesting to test. Furthermore, a study on the influence of the input feature's type would be interesting.
\item Rather than working in the pixel domain, DNNs may also be applied in the measurement domain for use in conjunction with compressive sensing data like in RPCA models \cite{RPCA-92,RPCA-17350-3}. 
\end{itemize}
Currently, only basic CNNs and GANs have been employed for background subtraction. Thus, future directions may investigate the adequacy and the use of pyramidal deep CNNs \cite{DNN-3100}, deep belief neural networks, deep restricted kernel neural networks \cite{DNN-4000}, probabilistic neural networks \cite{DNN-5000}, deep fuzzy neural networks \cite{DNN-6000,DNN-6001} and fully memristive neural networks \cite{DNN-7020,DNN-7000,DNN-7021,DNN-7011,DNN-7010,DNN-7015} in the case of static camera as well as moving camera \cite{Survey-10}. 

\bibliographystyle{plain}
\bibliography{mybib,rpca}
\end{document}